\pdfoutput=1
\documentclass[11pt]{article}

\usepackage[final]{acl}
\usepackage{times}
\usepackage{latexsym}
\usepackage{enumitem}
\definecolor{mypink}{rgb}{.99,.91,.95}
\definecolor{mygray}{rgb}{0.9, 0.9, 0.9}
\definecolor{linkblue}{rgb}{0.2,0.30,0.568}% 51, 77, 145
\newcommand{\hzz}[1]{{\color{red}{#1}}}

\definecolor{lightgreen}{rgb}{0.13, 0.55, 0.13}
\newcommand{\green}[1]{{\color{lightgreen}{#1}}}

\usepackage{hyperref}
\usepackage{url}
\usepackage{amssymb}
\usepackage{graphicx}
\usepackage{booktabs}
\usepackage{algorithm} 
\usepackage{algpseudocode}
\usepackage{mathrsfs}
\usepackage{multirow}
\usepackage{color}
\usepackage{colortbl}
\usepackage{arydshln}
\usepackage{array}
\usepackage{bm}
\usepackage{rotating}
\usepackage{adjustbox}
\usepackage{color}

\usepackage{textcomp}
\usepackage{mdframed}
\usepackage{amsmath}
\usepackage{CJKutf8}
\usepackage{algorithm} 
\usepackage{balance}
\usepackage{algpseudocode}

\usepackage[T1]{fontenc}
\usepackage[utf8]{inputenc}

\usepackage{microtype}

\usepackage{inconsolata}

\usepackage{graphicx}

\title{\raisebox{-1.0ex}{\includegraphics[width=0.08\textwidth]{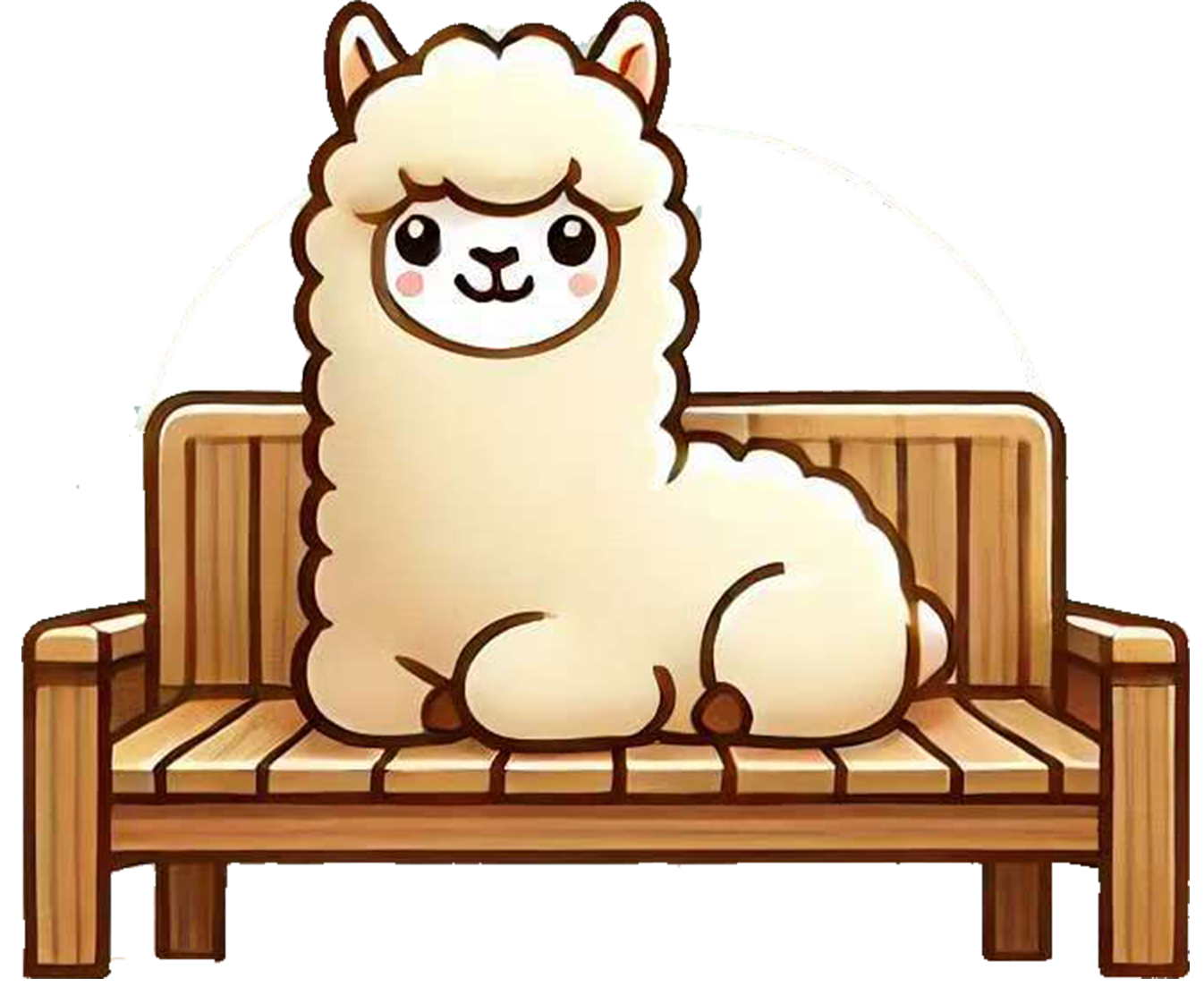}} MiniLongBench: The Low-cost Long Context Understanding Benchmark for Large Language Models}

\author{
 \textbf{Zhongzhan Huang\textsuperscript{1}},
 \textbf{Guoming Ling\textsuperscript{1}},
  \textbf{Shanshan Zhong\textsuperscript{1}},
  \textbf{Hefeng Wu\textsuperscript{1}},
   \textbf{Liang Lin\thanks{Corresponding author.}\textsuperscript{1,2,3}}
\\
 \textsuperscript{1}Sun Yat-sen University
 \textsuperscript{2}Peng Cheng Laboratory\\
 \textsuperscript{3}Guangdong Key Laboratory of Big Data Analysis and Processing
\\
}

\begin{document}
\maketitle
\begin{abstract}

Long Context Understanding (LCU) is a critical area for exploration in current large language models (LLMs). However, due to the inherently lengthy nature of long-text data, existing LCU benchmarks for LLMs often result in prohibitively high evaluation costs, like testing time and inference expenses. Through extensive experimentation, we discover that existing LCU benchmarks exhibit significant redundancy, which means the inefficiency in evaluation. In this paper, we propose a concise data compression method tailored for long-text data with sparse information characteristics. By pruning the well-known LCU benchmark LongBench, we create MiniLongBench. This benchmark includes only 237 test samples across six major task categories and 21 distinct tasks. Through empirical analysis of over 60 LLMs, MiniLongBench achieves an average evaluation cost reduced to only 4.5\% of the original while maintaining an average rank correlation coefficient of 0.97 with LongBench results. Therefore, our MiniLongBench, as a low-cost benchmark, holds great potential to substantially drive future research into the LCU capabilities of LLMs. See \href{https://github.com/MilkThink-Lab/MiniLongBench}{Github} for our code, data and tutorial. 
\end{abstract}

\section{Introduction}
\label{sec:intro}%

The ability for long context understanding, a.k.a LCU,~\citep{press2022train,sun2022length,chen2023extending,zeng2022glm,longchat2023,beltagy2020longformer,roy2021efficient} is one of the most important areas of exploration for current large language models (LLMs). Tasks with broad applications, such as summarization and question answering based on books, papers, and documents, as well as repository-level code generation, require the capability to handle long context sequences spanning thousands or even tens of thousands of tokens. 
\begin{figure}[t]
  \centering
\includegraphics[width=0.9\linewidth]{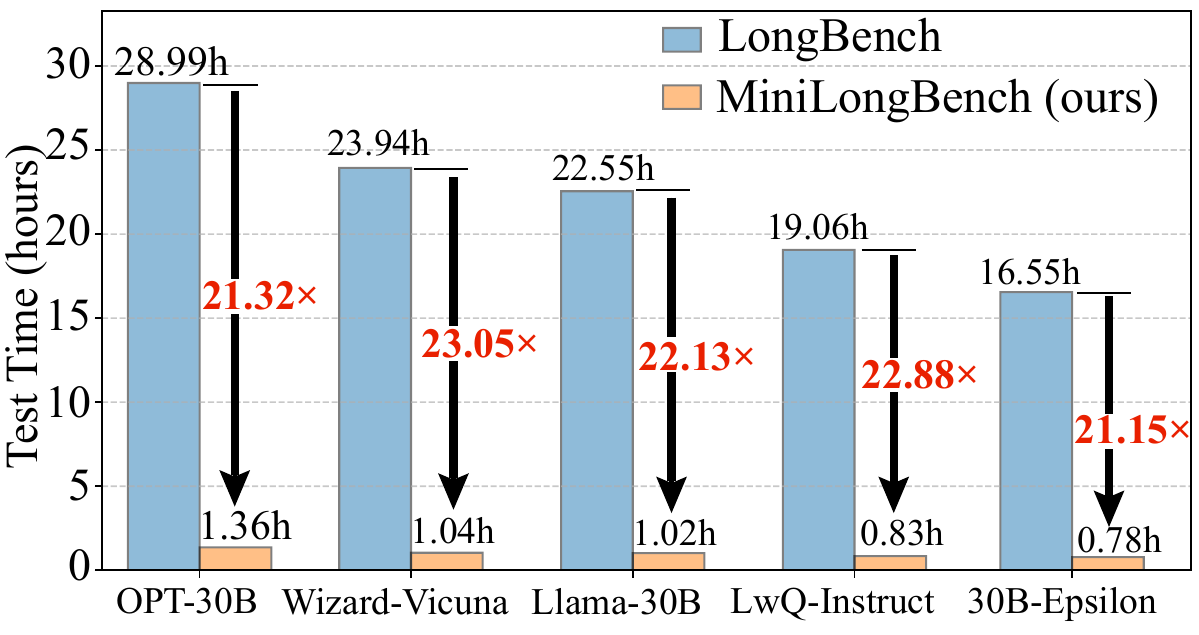}
\vspace{-10pt}
  \caption{The computational cost of LongBench and MiniLongBench. The proposed MiniLongBench effectively reduces the computational cost of the LongBench, thereby achieving a low-cost LCU benchmark.}
  \vspace{-15pt}
\label{fig:lowcost}
\end{figure}
\begin{figure*}[t]
  \centering
  \vspace{-5pt}
\includegraphics[width=0.99\linewidth]{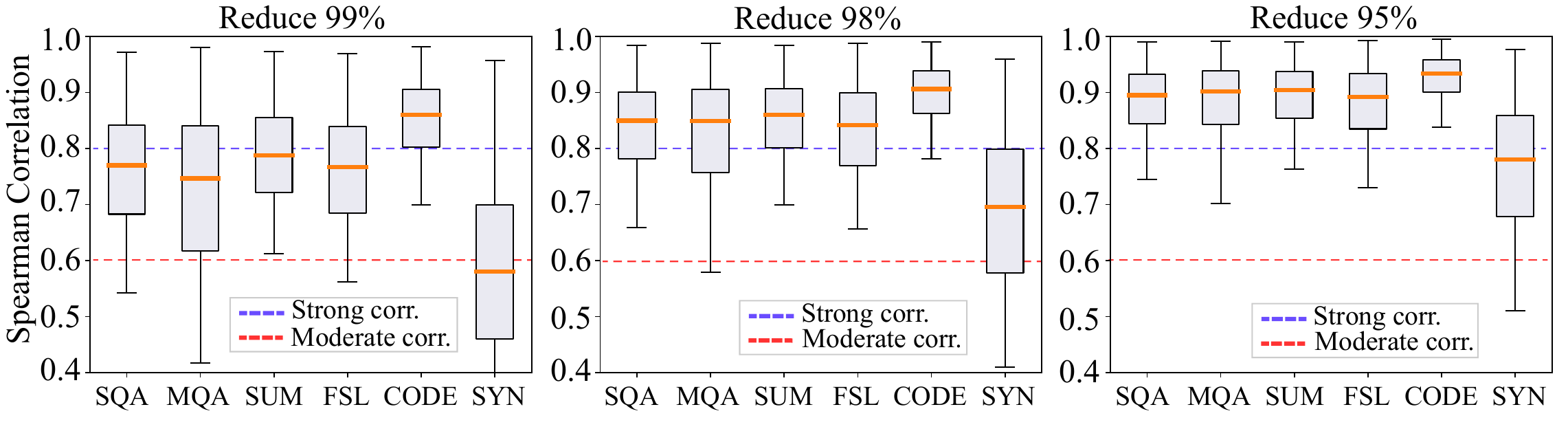}
\vspace{-10pt}
  \caption{The redundancy of LongBench. 
  "Reduce 95\%" means randomly removing 95\% of the dataset with equal probability. A Spearman correlation (Sp) $\geq$ 0.8 indicates a strong correlation and Sp $\geq$ 0.6 means moderate correlation between the results of randomly sampled subset and LongBench. The abscissa labels from "SQA" to "SYN" represent the abbreviations of the six main tasks in LongBench, with details provided in Section \ref{sec:redun}. }
  \vspace{-10pt}
\label{fig:redun}
\end{figure*}
Currently, the LCU capabilities of LLMs are still in their early stages, and their rapid development relies on recent proposals of LCU benchmarks~\citep{shaham2022scrolls,shaham2023zeroscrolls,an2023eval,bai2024LongBenchv2}. However, unlike normal LLM benchmarks~\citep{li2024seed,guo2023can,zhong2024let,zhong2023adapter}, LCU benchmarks inherently involve a large number of tokens due to the nature of long context data. Combined with the high number of test samples, the primary challenge these benchmarks face is their high evaluation cost. 
As shown in Fig.~\ref{fig:lowcost}, some popular LLMs on 8$\times$RTX3090 GPUs require approximately up to 15 $\sim$ 30 hours to complete an evaluation on LongBench with one batch size. 
Moreover, due to the large number of tokens in each long-text data, which significantly increases GPU memory consumption, it is challenging to accelerate testing through multi-batch processing. Therefore, the computational cost illustrated in Fig.~\ref{fig:lowcost} cannot be overlooked.
Furthermore, when researchers develop new LLM models and need to conduct multiple analyses of LCU capabilities, the time and computational costs become even more prohibitive. 
Given these challenges, we ask a critical question:

\begin{mdframed}[backgroundcolor=gray!8]
\begin{minipage}{\linewidth}
\vspace{2pt}
Do LCU benchmarks really need such a \textbf{large} number of test samples?
\vspace{2pt}
\end{minipage}
\end{mdframed}

To answer this question, in this paper, we explore the compression of the well-known LCU benchmark, LongBench~\citep{bai2023LongBench}. In Section \ref{sec:redun}, we first validate the significant redundancy in the LongBench through a series of random sampling experiments. Furthermore, in Section \ref{sec:method}, we propose a simple-yet-effective compression method for long-text data with  sparse information, resulting in a compact LCU benchmark, MiniLongBench. Finally, we explore the evaluation results of MiniLongBench across a range of existing LLMs. Our findings indicate that the proposed MiniLongBench substantially lowers the evaluation cost of LCU capabilities, reducing it to merely 4.5\% of the original, while maintaining the assessment outcomes of LLM on LongBench.
We show the related works in Appendix \ref{related}, and summarize the contributions of this paper as follows:
\begin{itemize}[leftmargin=15pt]
\vspace{-5pt}
    \item In this paper, we analyze the redundancy of current LCU benchmark for LLMs and propose an effective method to reduce the number of test samples for low-cost testing.
    \vspace{-3pt}
    \item Analyzing on over 60 LLMs, our MiniLongBench achieves an average ranking correlation of about 0.97 with LongBench while reducing computational cost to only 4.5\% of the original.
    \vspace{-15pt}
\end{itemize}

\begin{figure*}[t]
  \centering
  \vspace{-5pt}
\includegraphics[width=0.99\linewidth]{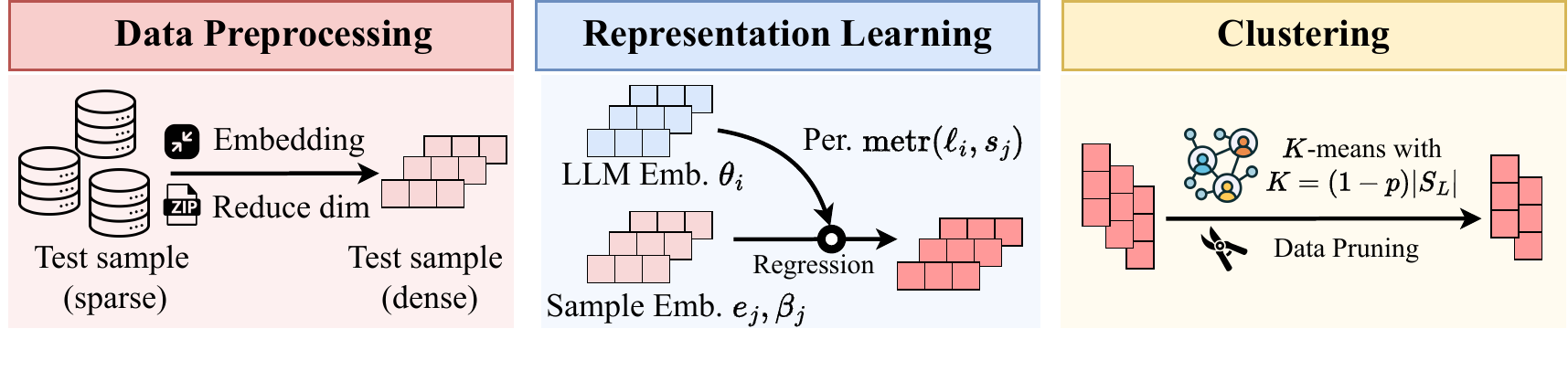}
\vspace{-10pt}
  \caption{The compression process of the LCU benchmark. "Emb." and "Per." respectively denote embedding and the performance of LLM $\ell_i$ on sample $s_j$ under the given metric $\text{metr}(\cdot, \cdot)$.}
  \vspace{-10pt}
\label{fig:alg}
\end{figure*}
\vspace{-5pt}

\section{The Redundancy of LCU Benchmark}
\label{sec:redun}

In this section, we consider the well-known LCU benchmark, LongBench~\citep{bai2023LongBench}, as an example to demonstrate that current LCU benchmarks suffer from significant redundancy. LongBench includes nearly 5000 test samples and covers six main task categories, such as  single-document question answering (SQA), multi-document question answering (MQA), summarization (SUM), few-shot learning (FSL), code completion (CODE), and synthetic tasks (SYN), which represent key long-text application scenarios. For specific details of LongBench, please refer to the Appendix~\ref{app:LongBench}.

First, we randomly sample the long-text data from different categories of LongBench $n$=10,000 times to obtain $n$ subsets of test samples with compression ratio $p$, where $p$ represents the proportion of remaining samples after sampling. 
We then test these subsets using dozens of LLMs (see the Appendix \ref{app:llms} for details), and compute the Spearman correlation (Sp) coefficient \citep{spearman1961proof} to measure the ranking correlation between the evaluation results of each subset and those of the original LongBench $S_L$. The closer "Sp" is to 1.0, the more the evaluation of the sampled subset aligns with the evaluation of $S_L$. We take $p \in \{0.99, 0.98, 0.95\}$ and select the top 7500 results based on Sp for statistical analysis. The experimental results are shown in Fig~\ref{fig:redun}.
We find that even though LongBench data is randomly reduced by a large amount, some subsets of LongBench still exhibit strong ranking correlations with the original benchmark, e.g., with Sp greater than 0.8 and even 0.9. This indicates that LongBench contains significant redundancy and \textbf{does not require so many test samples}. Therefore, in this paper, we will design an efficient method to create a more compact LCU benchmark.

\section{Compression for LCU Benchmark}
\label{sec:method}

In this section, we explore how to filter long-text data to reduce the size of LCU benchmark, enabling our MiniLongBench for low-cost estimation of LCU capabilities. 
Although Fig.~\ref{fig:redun} shows that random sampling can also yield subsets with large Sp, due to its high variance, we need to develop a more stable compression method.
A straightforward intuition is that, given a set of $m$ LLMs $\{\ell_i\}_{i=1}^m$, we can leverage their performance on all test samples from LongBench $S_L = \{s_j\}_{j=1}^{|S_L|}$.
Using their performance record, we aim to construct a regression model to map these sparsely informative long-text samples into a denser text space first, and gradually project them into a performance space. After that, we can learn the representation of the test samples.    
Moreover, we then cluster these samples and retain only a certain number of cluster centers as representative test samples, forming $S_{\text{mini}}$, thereby compressing the benchmark. Fig.~\ref{fig:alg} and Alg.~\ref{alg:11} show the compression process of the LCU benchmark and construction of proposed MiniLongBench. Specifically,

\begin{algorithm}[tb]
\small
   \caption{The construction of MiniLongBench.}
    \textbf{Input:}  The long-text data $S_L = (s_1, s_2,...,s_{|S_L|})$, reduced dimension $d$ and ratio $p$. The performance record $\text{metr}(\ell_i,s_j)$ from LLM $\ell_i$ and sample $s_j$.  Text embedding $\kappa_{\text{test}}$. 
    
    \textbf{Output:} Compact LCU benchmark.
   % \vspace{-15pt}
\begin{algorithmic}

\State\green{\algorithmiccomment{Data preprocessing}}

\State Intra-sample dimension reduction $s_j^\prime \gets \kappa_{\text{text}}(s_j)$;
\State Inter-sample dimension reduction by 
\State\quad $\{e_j\}_{j=1}^{|S_L|}  \gets \text{PCA}_d[s_1^\prime, s_2^\prime, ..., s_{|S_L|}^\prime]$;

    \State\green{\algorithmiccomment{Representation learning for test samples}}

\State Initialize LLM $\ell_i$'s representation by $\theta_i \sim \mathcal{N}(\mathbf{0},\mathbf{I}_d)$;
\State Initialized $\beta_j \gets \mathbf{0}$;
\State Update learnable $(e_j,\beta_j)$ and $\theta_i$ by Eq.~(\ref{eq:logi});

    \State\green{\algorithmiccomment{Clustering}}
\State Determine the number of cluster centers $K \gets (1-p)|S_L|$;
\State Obtain $K$ centers $(c_1,c_2,..,c_K)$ by clustering and $(e_j,\beta_j)$; 
\State $S_{\text{mini}} \gets (c_1,c_2,..,c_K)$;

\State\Return Compact LCU benchmark $S_{\text{mini}}$
% \vspace{-15pt}
\end{algorithmic}
\label{alg:11}

\end{algorithm}
% \vspace{-15pt}

% Table generated by Excel2LaTeX from sheet 'Sheet1'
\begin{table*}[t]
  \centering
  \resizebox*{0.99\linewidth}{!}{
    \begin{tabular}{llllrrrr}
    \toprule
    \textbf{Dataset}  & \textbf{Index} & \textbf{Metric} & \textbf{Language} & \multicolumn{1}{l}{\textbf{Long. Avg len}} & \multicolumn{1}{l}{\textbf{MiniLong. Avg len}} & \multicolumn{1}{l}{\textbf{Long. \#data}} & \multicolumn{1}{l}{\textbf{MiniLong. \#data}} \\
    \midrule
    \rowcolor{mypink}\textbf{Single-Document QA} & &       &       &       &       &       &  \\
    NarrativeQA & 1-1& F1    & English & 18,409 & 22,967 & 200   & 6 (\hzz{$\downarrow$ 97\%})\\
    Qasper & 1-2& F1    & English & 3,619 & 2,933 & 200   & 9 (\hzz{$\downarrow$ 96\%})\\
    MultiFieldQA-en & 1-3& F1    & English & 4,559 & 4,519 & 150   & 7 (\hzz{$\downarrow$ 95\%})\\
    MultiFieldQA-zh & 1-4& F1    & Chinese & 6,701 & 6,300 & 200   & 15 (\hzz{$\downarrow$ 93\%})\\
    \midrule
    \rowcolor{mypink}\textbf{Multi-Document QA} & &      &       &       &       &       &  \\
    HotpotQA & 2-1& F1    & English & 9,151 & 8,856 & 200   & 13 (\hzz{$\downarrow$ 94\%})\\
    2WikiMultihopQA & 2-2& F1    & English & 4,887 & 4,286 & 200   & 13 (\hzz{$\downarrow$ 94\%})\\
    MuSiQue & 2-3& F1    & English & 11,214 & 10,910 & 200   & 7 (\hzz{$\downarrow$ 97\%})\\
    DuReader & 2-4& Rouge-L & Chinese & 15,768 & 12,996 & 200   & 6 (\hzz{$\downarrow$ 97\%})\\
    \midrule
    \rowcolor{mypink}\textbf{Summarization} & &       &       &       &       &       &  \\
    GovReport & 3-1& Rouge-L & English & 8,734 & 7592 & 200   & 12 (\hzz{$\downarrow$ 94\%})\\
    QMSum & 3-2& Rouge-L & English & 10,614 & 8,253 & 200   & 6 (\hzz{$\downarrow$ 97\%})\\
    MultiNews & 3-3& Rouge-L & English & 2,113 & 1,785 & 200   & 11 (\hzz{$\downarrow$ 95\%})\\
    VCSUM & 3-4& Rouge-L & Chinese & 15,380 & 10,400 & 200   & 6 (\hzz{$\downarrow$ 97\%})\\
    \midrule
    \rowcolor{mypink}\textbf{Few-shot Learning} & &       &       &       &       &       &  \\
    TREC  & 4-1& Acc. (CLS) & English & 5,177 & 6,077 & 200   & 8 (\hzz{$\downarrow$ 96\%})\\
    TriviaQA & 4-2& F1    & English & 8,209 & 9,719 & 200   & 12 (\hzz{$\downarrow$ 94\%})\\
    SAMSum & 4-3& Rouge-L & English & 6,258 & 5,974 & 200   & 15 (\hzz{$\downarrow$ 93\%})\\
    LSHT  & 4-4& Acc. (CLS) & Chinese & 22,337 & 22,759 & 200   & 8 (\hzz{$\downarrow$ 96\%})\\
    \midrule
    \rowcolor{mypink}\textbf{Synthetic Task} & &       &       &       &       &       &  \\
    PassageCount & 5-1& Acc. (EM) & English & 11,414 & 10,627 & 200   & 4 (\hzz{$\downarrow$ 98\%})\\
    PassageRetrieval-en & 5-2& Acc. (EM) & English & 9,289 & 9,394 & 200   & 15 (\hzz{$\downarrow$ 93\%})\\
    PassageRetrieval-zh & 5-3& Acc. (EM) & Chinese & 6,745 & 6,684 & 200   & 15 (\hzz{$\downarrow$ 93\%})\\
    \midrule
    \rowcolor{mypink}\textbf{Code Completion} & &       &       &       &       &       &  \\
    LCC   & 6-1& Edit Sim & Python/C\#/Java & 1,235 & 1,187 & 500   & 26 (\hzz{$\downarrow$ 95\%})\\
    RepoBench-P & 6-2& Edit Sim & Python/Jave & 4,206 & 3,723 & 500   & 23  (\hzz{$\downarrow$ 95\%}) \\
    \bottomrule
    \end{tabular}%
    }
\caption{The dataset statistics in LongBench and MiniLongBench.  "Long." and "MiniLong." denote LongBench and MiniLongBench. "Avg len" (average length) is computed using the number of words for the English (code) datasets and the number of characters for the Chinese datasets. "Acc. (CLS)" refers to classification accuracy, while "Acc. (EM)" refers to exact match accuracy. "\#data" means the number of data.}
\label{tb:stat}
\end{table*}%

\noindent(1) \underline{\textbf{Data Preprocessing}}.  Unlike data from conventional LLM benchmarks, the effective information in long-text data is highly sparse. Without proper compression of this information, it can significantly impact subsequent representation learning and clustering processes. 
Therefore, for the sparse information in these long-text data, we initially densify them using a text encoder $\kappa_{\text{text}}$ OpenAIEmbedding~\citep{xian2024vector} and a principal component analysis, a.k.a PCA,~\citep{abdi2010principal} to obtain part of dense $d-$dimentional initialization of test samples, i.e., 
\vspace{-3pt}
% \begin{equation}
%     e_j = \kappa_{\text{text}} (s_j),\quad j = 1,2, ..., |S_L|.
% \end{equation}
\begin{equation}
    \{e_j\}_{j=1}^{|S_L|}  = \text{PCA}_d[\{\kappa_{\text{text}} (s_j)\}_{j=1}^{|S_L|}],
    \label{eq:pca}
    \vspace{-2pt}
\end{equation}
For a detailed discussion on how data preprocessing influences the construction of MiniLongBench, please refer to Section \ref{sec:ana}.

\noindent(2) \underline{\textbf{Representation Learning}}. 
Moreover, similar to \cite{polotinybenchmarks}, which utilizes the Item Response Theory in psychology and education~\citep{cai2016item}, we can perform representation learning for test samples. Suppose we have LLMs $\{\ell_i\}_{i=1}^m$ and test samples $s_j \in S_L$ with performance measured by the metric $\text{metr}(\cdot, \cdot)$. We then assume that the probability of LLM $\ell_i$ correctly answering sample $s_j$ is given by:
\vspace{-10pt}
\begin{multline}
    \mathbb{P}(\text{metr}(\ell_i, s_j)=1|\theta_i,e_j,\beta_j) \\
    = [1+\exp(-e_j\top \theta_i + \beta_j)]^{-1},
    \label{eq:logi}
    \vspace{-5pt}
\end{multline}
where the learnable parameter $\theta_i \in \mathbb{R}^d$ represents the $d$-dimensional embedding of LLM $\ell_i$, initialized using a $d$-dimensional standard normal distribution. Eq.~(\ref{eq:logi}) is classical logistic regression model~\citep{kleinbaum2002logistic}. The parameters $(e_j, \beta_j)$ are the learnable representations of the test sample $s_j$, where $\beta_j$ is initialized to zero vector and $e_j$ initialized by Eq.~(\ref{eq:pca}). In this paper, we set $d=10$. See further analysis on these representations, initialization and $d$ in Section \ref{sec:ana}.

It is worth noting that in Eq.~(\ref{eq:logi}), we use a binary classification example for the metric, where $\text{metr}(\ell_i, s_j) = 1$ if $\ell_i$ performs correctly on $s_j$, and $\text{metr}(\ell_i, s_j) = 0$ otherwise. If $\text{metr}(\cdot, \cdot)$ is continuous metrics, it can also be transformed into a binary classification scenario.
Specifically, the metric 
$\text{metr}(\cdot, \cdot)$
 is generally bounded. For example, in LongBench, metrics such as F1 score, edit distance, etc., are used. We can normalize them to the interval $[0,1]$
, and then consider the following optimization problem.
\begin{multline}
    \min_c \|\sum\nolimits_{i=1}^m\sum\nolimits_{j=1}^{|S_L|} \text{metr}(\ell_i, s_j) \\- \sum\nolimits_{i=1}^m\sum\nolimits_{j=1}^{|S_L|} \mathbf{1}_{[\text{metr}(\ell_i, s_j) \geq c ]}\|,
\label{eq:min}
\end{multline}
Note that the existence of a solution to Eq.~(\ref{eq:min}) is evident. It can be obtained by simply searching the interval $[0,1]$
 to get an approximate solution for $c$. Once $c$
 is obtained, we replace the original $\text{metr}(\ell_i, s_j)$
 with $\text{metr}(\ell_i, s_j)^\prime = \mathbf{1}_{[\text{metr}(\ell_i, s_j) \geq c ]}$
which can transform the continuous metric into a discrete binary scenario similar to Eq.~(\ref{eq:logi}).

\begin{figure}[t]
  \centering
  \vspace{-5pt}
\includegraphics[width=0.99\linewidth]{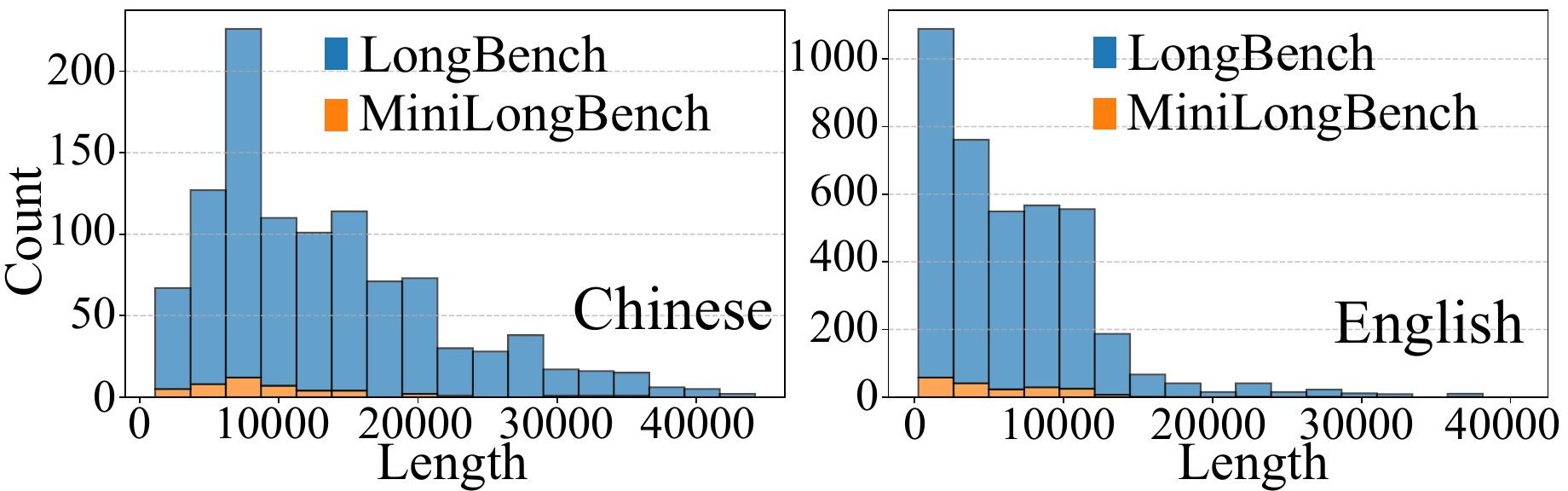}
\vspace{-10pt}
  \caption{The Length distribution for English and Chinese data in LongBench and MiniLongBench, measured by the number of words and characters.}
  \vspace{-1pt}
\label{fig:count}
\end{figure}
\noindent(3) \underline{\textbf{Clustering}}. 
Next, we update $\theta_i$, $e_j$, and $\beta_j$ simultaneously using the training approach of logistic regression. Once these learnable parameters converge, we concatenate $(e_j, \beta_j)$ as the final representation of the test sample $s_j$, and perform clustering analysis on them using K-Means~\citep{hamerly2003learning} under Euclidean distance, where $K=(1-p)|S_L|$. Finally, the all cluster centers are integrated as the test samples of MiniLongBench $S_{\text{mini}}$. In Section \ref{sec:MiniLongBench}, we will further validate the effectiveness of MiniLongBench from an experimental perspective.

\section{Compact LCU Benchmark: MiniLongBench}
\label{sec:MiniLongBench}

In this section, we present our compact MiniLongBench and demonstrate through comprehensive experiments that it significantly reduces computational costs while preserving original LongBench's evaluation effectiveness. We select over 60 LLMs for analysis, with 
$m=20$
 of them participating in the training described in Section \ref{sec:method}, and the rest serving as candidates for validating the effectiveness of MiniLongBench. See Appendix \ref{app:llms} for details of LLMs considered.

\begin{table}[t]
  \centering
  
  \resizebox*{0.99\linewidth}{!}{
    \begin{tabular}{l|cccccc}
    \toprule
    \textbf{Model} & \textbf{SQA}   & \textbf{MQA}   & \textbf{SUM}   & \textbf{FSL}   & \textbf{SYN}   & \textbf{CODE} \\
    \midrule
    DeepSeek-V3-128k & 0.43  & 0.31  & 0.12  & 0.67  & 0.26  & 0.88  \\
    \rowcolor{mygray}GPT-4o-mini-128k & 0.41  & 0.30  & 0.10  & 0.65  & 0.23  & 0.87  \\
    GPT-3.5-Turbo-16k & 0.37  & 0.25  & 0.07  & 0.58  & 0.17  & 0.81  \\
    \rowcolor{mygray}Internlm3-8B-32k & 0.33  & 0.24  & 0.04  & 0.53  & 0.12  & 0.64  \\
    ChatGLM3-6B-8k & 0.17  & 0.08  & 0.02  & 0.39  & 0.02  & 0.48  \\
    \rowcolor{mygray}ChatGLM4-9B-128k & 0.36  & 0.27  & 0.06  & 0.58  & 0.14  & 0.75  \\
    Qwen-7B-8k & 0.20  & 0.10  & 0.04  & 0.48  & 0.07  & 0.71  \\
    \rowcolor{mygray}Qwen2-7B-128k & 0.29  & 0.22  & 0.05  & 0.55  & 0.09  & 0.71  \\
    Qwen2.5-7B-128k & 0.34  & 0.24  & 0.04  & 0.55  & 0.12  & 0.67  \\
    \rowcolor{mygray}Qwen2.5-14B-128k & 0.35  & 0.26  & 0.05  & 0.56  & 0.12  & 0.66  \\
    Qwen2.5-32B-128k & 0.36  & 0.26  & 0.06  & 0.59  & 0.15  & 0.74  \\
    \rowcolor{mygray}Llama-7B-2k & 0.09  & 0.04  & 0.02  & 0.39  & 0.02  & 0.61  \\
    Llama2-7B-4k & 0.11  & 0.04  & 0.02  & 0.41  & 0.04  & 0.66  \\
    \rowcolor{mygray}Llama3-8B-8k & 0.11  & 0.03  & 0.04  & 0.47  & 0.08  & 0.71  \\
    Llama-30B-2k & 0.10  & 0.04  & 0.02  & 0.41  & 0.04  & 0.65  \\
    \rowcolor{mygray}OPT-30B-2k & 0.08  & 0.03  & 0.01  & 0.33  & 0.02  & 0.48  \\
    Wizard-Vicuna-2k & 0.20  & 0.13  & 0.02  & 0.41  & 0.05  & 0.59  \\
    \rowcolor{mygray}LwQ-Instruct-2k & 0.23  & 0.18  & 0.04  & 0.45  & 0.07  & 0.70  \\
    30B-Epsilon-2k & 0.20  & 0.13  & 0.04  & 0.48  & 0.07  & 0.73  \\
    \bottomrule
    \end{tabular}%
    }
    \caption{Specific evaluation results on MiniLongBench. See Appendix \ref{app:fur} and Appendix \ref{app:dire} for the more analysis and detail results on various advanced LLMs.  }
    \vspace{-5pt}
  \label{tab:evalstat}%
\end{table}%

\subsection{The Details of MiniLongBench}
\label{app:MiniLongBench}
Chosing compression ratio $p=0.95$, we use the compression method shown in Section~\ref{sec:method} for LongBench to obtain compact LCU benchmark MiniLongBench. 
This benchmark includes only 237 test samples across six task categories, with an average length of 6193 words (English) and 10344 characters (Chinese). Consistent with LongBench, MiniLongBench has six major task categories and 21 distinct tasks, covering key long-text application scenarios. 
Through the long-text dataset compression method proposed in Alg.~\ref{alg:11}, these different tasks have been compressed by about 95\%, greatly reducing the computational consumption of the LCU benchmark in the testing process. The specific statistics is shown in Table~\ref{tb:stat}.

As shown in Table \ref{tb:stat}, the average length of MiniLongBench is smaller compared to that of LongBench due to data pruning, but it generally maintains a similar magnitude. This indicates that LongBench retains a good diversity of long-text data even after compression. 
Moreover, we further illustrate the length distribution of data across different languages, including English and Chinese, in Fig.~\ref{fig:count}. We observe that for different languages, our proposed MiniLongBench significantly reduces the total length of data input to the LLM, thereby greatly decreasing the number of tokens in the model input and reducing computational costs. The further discussions with other compression ratio $p$ and $m$ trained LLMs are shown in Section \ref{sec:ana}.

\begin{figure}[t]
  \centering
  \vspace{-5pt}
\includegraphics[width=0.99\linewidth]{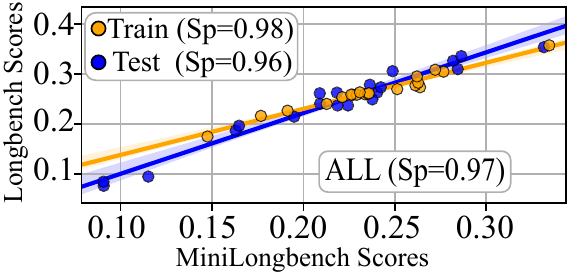}
\vspace{-10pt}
  \caption{The analysis of rank correlation (Sp) between LongBench and MiniLongBench.}
  \vspace{-10pt}
\label{fig:spall}
\end{figure}

\subsection{The Evaluation Method}
\label{app:perfo2}

In this section, we explore how to evaluate the LCU capabilities of LLMs using MiniLongBench. A straightforward approach is to directly assess them on MiniLongBench, yielding reliable results with a Sp of 0.95 compared to LongBench (see Appendix \ref{app:dire} for details). However, it's important to note that MiniLongBench, having significantly fewer test samples than LongBench, may introduce some evaluation bias. To mitigate this, we can use MiniLongBench samples to estimate the performance~\cite{polotinybenchmarks,pacchiardi2024100} of the LLMs on LongBench, thereby reducing bias and achieving an improved Sp of up to 0.97.

Specifically, For a new LLM $\ell_0$ to be tested, we first evaluate it on all test samples $c_j$ from MiniLongBench $S_{\text{mini}}$ and obtain its performance $\text{metr}(\ell_0, c_j)$. Subsequently, we apply consistent normalization and discretization for $\text{metr}(\ell_0, c_j)$ as outlined in Section \ref{sec:method}, and initialize a $d$-dimensional feature vector $\bar{\theta}$ for the LLM $\ell_0$ using a standard normal distribution.

Next, we fine-tune $\bar{\theta}$ on the test samples of $S_{\text{mini}}$ using Eq.~(\ref{eq:logi}) to adapt it to the representation space of the test samples. After completing the fine-tuning, we can construct the following MiniLongBench score through Eq.~(\ref{eq:logi}) to estimate the performance of $\ell_0$ across the entire $S_L$:
\vspace{-1pt}
\begin{equation}
\vspace{-1pt}
 \sum\nolimits_{j=1}^{|S_L|} [1+\exp(-e_j\top \bar{\theta} + \beta_j)]^{-1}/ |S_L|,
    \label{eq:score}
\end{equation}
The time required for fine-tuning in the aforementioned evaluation process and the storage cost for the features of $S_L$ are both minimal, requiring only about 10 MB of disk space and as little as 0.03 seconds of GPU time, even on a laptop. For specific statistics, please refer to Appendix \ref{app:cost}.

\subsection{The Evaluation Results}
\label{app:perfo}
 
Moreover, Fig.~\ref{fig:spall} shows the rank correlation between LongBench and the proposed MiniLongBench are 0.96$\sim$0.98, whether on the LLMs that participated in the training or on other unseen LLMs. Moreover, in conjunction with the results presented in Fig.\ref{fig:lowcost}, this indicates that the proposed MiniLongBench can effectively replicate the evaluation outcomes of LongBench while maintaining very low computational costs. 

Additionally, we present in Table \ref{tab:evalstat} the specific performance of various advanced LLMs across different tasks on the proposed MiniLongBench. For more detailed results, please refer to Appendix~\ref{app:fur}.

\begin{figure}[t]
  \centering
  \vspace{-1pt}
\includegraphics[width=0.99\linewidth]{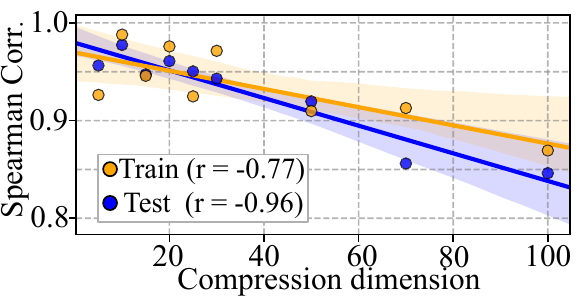}
\vspace{-15pt}
  \caption{The impact of compression dimension $d$ on the construction of MiniLongBench. "r" is Pearson correlation coefficient.}
  \vspace{-1pt}
\label{fig:d}
\end{figure}

\section{Analysis}
\label{sec:ana}
In this Section, We conduct a more comprehensive analysis of the proposed MiniLongBench.

\begin{mdframed}[backgroundcolor=gray!8]
\begin{minipage}{\linewidth}
\vspace{2pt}
\noindent\textbf{(1) How does the reduced dimension $d$ affect the compression of the LCU benchmark?}
\vspace{2pt}
\end{minipage}
\end{mdframed}
In Eq.~(\ref{eq:pca}) of Session \ref{sec:method}, we perform initial compression of the long-text data in LongBench using text embedding OpenAIEmbedding~\citep{xian2024vector} and a PCA~\citep{abdi2010principal}, allowing the long-text information to be initialized as some vectors with dimension $d$. 

In this section, we further explore the specific impact of the compressed dimension $d$ on constructing a compact MiniLongBench. Specifically, following the experiment setting in Section \ref{sec:MiniLongBench}, we consider $d \in \{5, 10, 15, 20, 25, 30, 50, 70, 100\}$ and present the Sp of the evaluation results for LongBench and MiniLongBench under different values of $d$ in Fig.~\ref{fig:d}. We observe that a negative correlation between Sp and $d$. This indicates that for long-texts data with sparse information, using excessively high-dimensional representations is not advisable, as it can still lead to sparse representations even after representation learning. Further information compression is crucial. In this paper, we set $d=10$ by default.
\begin{figure}[t]
  \centering
  \vspace{-1pt}
\includegraphics[width=0.95\linewidth]{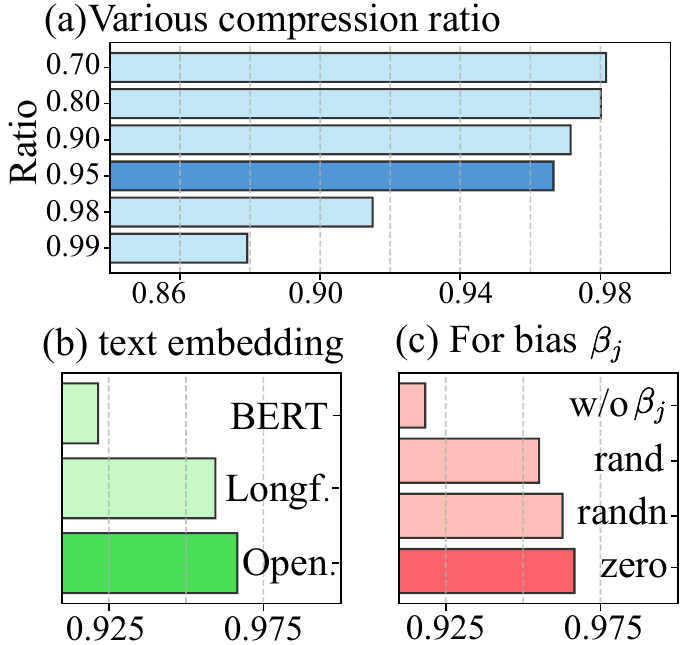}
\vspace{-5pt}
  \caption{Further analysis for MiniLongBench. The results of (a) various compression ratio $p$, (b) various text embedding $\kappa_{\text{text}}$. (c) The influence of various $\beta_j$. The bars with darker color represent the settings adopted by our settings. "rand" and "randn" denote the standard uniform and normal distribution. "Longf." and "Open." are Longformer and OpenAIEmbedding.}
  \vspace{-1pt}
\label{fig:abl}
\end{figure}
\begin{figure*}[t]
  \centering
  \vspace{-5pt}
\includegraphics[width=0.99\linewidth]{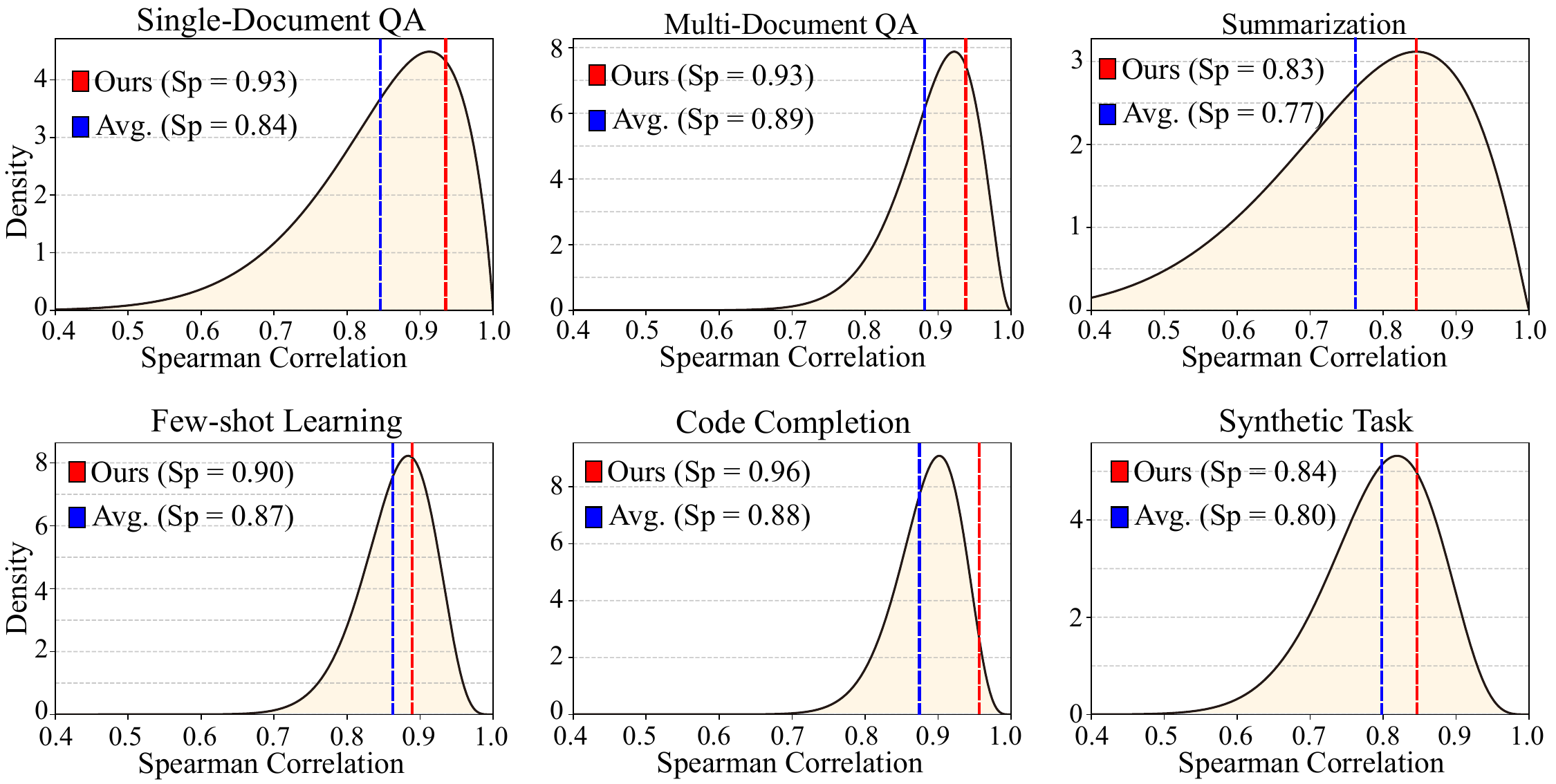}
\vspace{-5pt}
  \caption{The impact of the selection of LLMs on the construction of MiniLongBench .}
  % \vspace{-10pt}
\label{fig:select}
\end{figure*}

\begin{mdframed}[backgroundcolor=gray!8]
\begin{minipage}{\linewidth}
\vspace{2pt}
\noindent\textbf{(2) Is PCA necessary for MiniLongBench?}
\vspace{2pt}
\end{minipage}
\end{mdframed}
In data preprocessing, PCA is employed to further reduce the dimensionality of features after text embedding. If we remove the PCA operation in Eq.~(\ref{eq:pca}), on one hand, the high dimension of $\kappa_{\text{text}}$, which reaches 1024, would result in significant additional computational overhead during training. Moreover, due to the large dimensionality, we observe that the average Sp of MiniLongBench and LongBench drops from 0.95 to 0.67, which aligns with the phenomenon observed in Fig.~\ref{fig:d}. Therefore, the dimension reducing method, like PCA, is essential for the construction of MiniLongBench.

\begin{mdframed}[backgroundcolor=gray!8]
\begin{minipage}{\linewidth}
\vspace{2pt}
\noindent\textbf{(3) How to select the text embedding $\kappa_{\text{text}}$}.
\vspace{2pt}
\end{minipage}
\end{mdframed}
In the data preprocessing phase, we utilize OpenAIEmbedding~\citep{xian2024vector} for text embedding. In Fig.~\ref{fig:abl} (b), we present the results of employing alternative text embeddings, including Longformer~\cite{zhu2021long} and BERT~\cite{Liu2019RoBERTaAR}. We observe that BERT, which only supports token inputs with a maximum length of 512, significantly underperforms compared to OpenAIEmbedding and Longformer, which support lengths of 8192 and 4096, respectively. This is primarily due to BERT's weaker capability in information densification and the inevitable information loss when handling test samples exceeding the token length limit, as they can only be processed through chunked densification. Therefore, this paper defaults to using the more capable OpenAIEmbedding.

\begin{mdframed}[backgroundcolor=gray!8]
\begin{minipage}{\linewidth}
\vspace{2pt}
\noindent\textbf{(4) How about other compression ratios $p$?}
\vspace{2pt}
\end{minipage}
\end{mdframed}
In this paper, we set the compression ratio $p=0.95$ as the default. Subsequently, we further explore the selection of $p$ in Fig.~\ref{fig:abl} (a). We observe that as $p$ approaches 1, meaning more test samples are reduced, the Sp between MiniLongBench and LongBench decreases, which aligns with the observations in Fig.~\ref{fig:redun}. This is because, although the LCU benchmark has significant redundancy in test samples, an extremely low compression ratio can easily disrupt the data distribution or diversity of the benchmark, leading to substantial bias in the evaluation of LLMs. Based on the experimental results in Fig.~\ref{fig:abl} (a), $p=0.95$ is a favorable choice, as it balances both the testing cost and the evaluation capability of the benchmark.

\begin{mdframed}[backgroundcolor=gray!8]
\begin{minipage}{\linewidth}
\vspace{2pt}
\noindent\textbf{(5) Is the learnable bias $\beta_j$ important?}
\vspace{2pt}
\end{minipage}
\end{mdframed}
In Eq.~(\ref{eq:logi}), we introduce a learnable bias $\beta_j$ for the logistic regression model. In Fig.~\ref{fig:abl} (c), we explore its impact on the construction of MiniLongBench by testing different initializations and removing it entirely. We observe that, on one hand, the inclusion of $\beta_j$ aids in the representation learning of test samples, as removing it results in a noticeable decline in Sp. On the other hand, different initializations yield varying performance levels, with zero initialization achieving the best results. In conclusion, the setting of learnable $\beta_j$ is important, and we employ a learnable bias with zero initialization.

\begin{mdframed}[backgroundcolor=gray!8]
\begin{minipage}{\linewidth}
\vspace{2pt}
\noindent\textbf{(6) About the selection of $m$ LLMs.}
\vspace{2pt}
\end{minipage}
\end{mdframed}
In this section, we further analyze the impact of the LLMs involved in training on the construction of MiniLongBench from the selection of LLMs.

We fix the number of LLMs, $m$, and then independently sample 1000 times from all the LLMs considered in this paper. Using the method mentioned in Section \ref{sec:method}, we obtain various compact new "MiniLongBench" and compute its Sp distribution against LongBench evaluation results. The results are shown in Fig.~\ref{fig:select}. We find that the choice of LLMs involved in training significantly affects the construction of MiniLongBench, which is intuitive. This is because the representation of test samples depends on the performance records of the LLMs on LongBench, and when the selected LLMs perform poorly, their representations struggle to correctly project the test samples into the performance space. In this paper, we manually select a few LLMs with generally good performance across various aspects to participate in the construction of MiniLongBench. A list of the chosen LLMs can be found in Appendix \ref{app:llms}. In the future, the automated LLMs selection is needed.

\begin{figure*}[t]
  \centering
  \vspace{-5pt}
\includegraphics[width=0.99\linewidth]{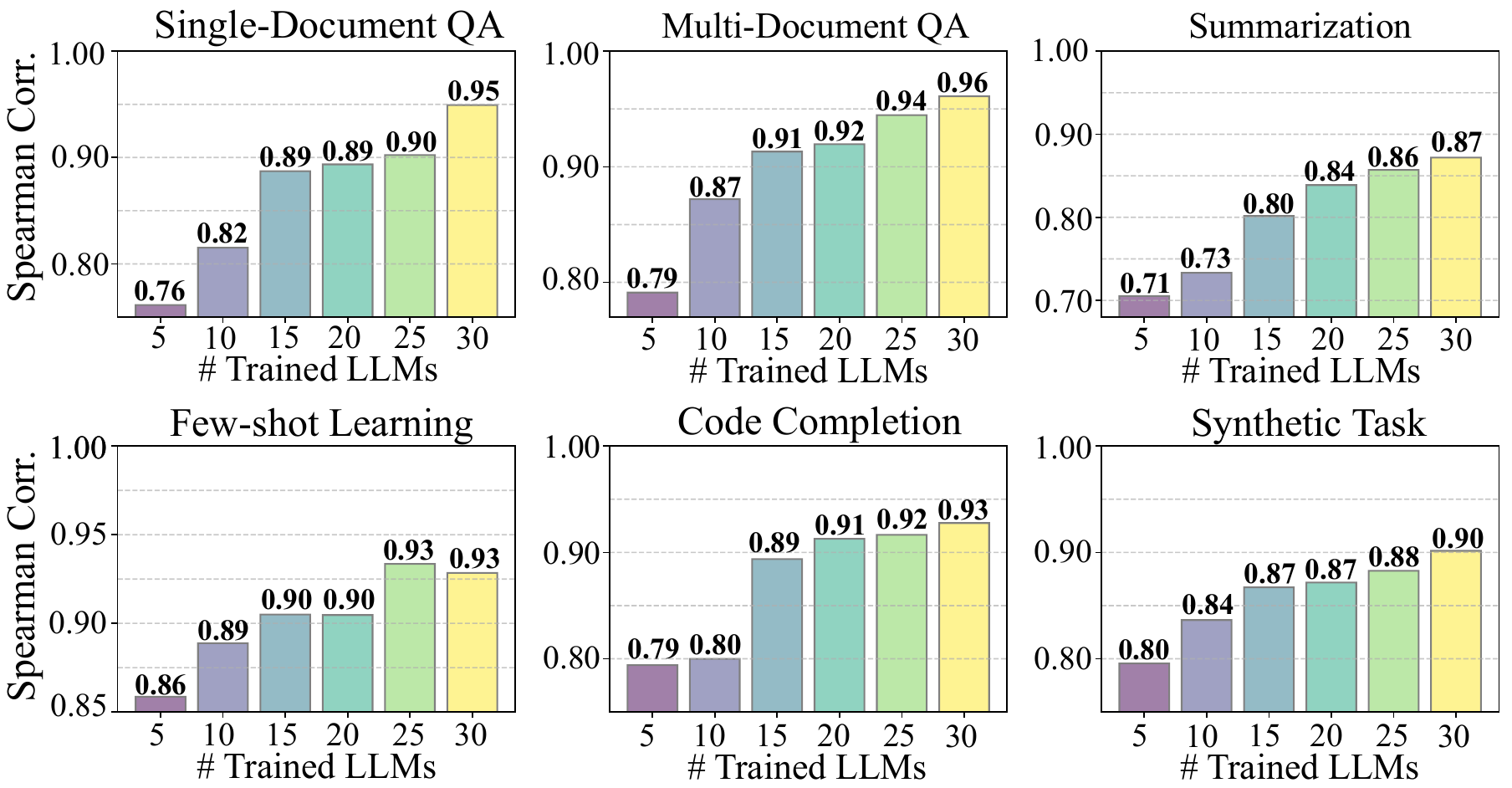}
\vspace{-3pt}
  \caption{The impact of the number of LLMs $m$ on the construction of MiniLongBench .}
  % \vspace{-10pt}
\label{fig:m}
\end{figure*}

\begin{mdframed}[backgroundcolor=gray!8]
\begin{minipage}{\linewidth}
\vspace{2pt}
\noindent\textbf{(7) What is the appropriate number of LLMs $m$ to involve in training? }
\vspace{2pt}
\end{minipage}
\end{mdframed}
Moreover, we further explore the impact of the number of LLMs involved in training. For a specific number of LLMs, $m$, we repeat the independent sampling 5 times and compute the average Sp of the constructed MiniLongBench and LongBench evaluation results across all LLMs. The results are shown in Fig.~\ref{fig:m}. We observe that as $m$ increases, Sp gradually increases and approaches 1.0. This indicates that involving enough LLMs is beneficial for the representation learning of test samples. 
And we also note that when $m = 20$, the Sp in different tasks seems acceptable, suggesting that although the number of LLMs aids in representation learning, there is still considerable redundancy. Considering the computational cost, we take the acceptable $m=20$ as default. 

\begin{mdframed}[backgroundcolor=gray!8]
\begin{minipage}{\linewidth}
\vspace{2pt}
\noindent\textbf{(8) Is the average Sp $\geq$ 0.97 enough? }
\vspace{2pt}
\end{minipage}
\end{mdframed}
In Section \ref{sec:MiniLongBench}, we show that the proposed MiniLongBench achieves an average Sp of 0.97 compared to LongBench. And, we also find that the p-value is less than 0.001, indicating the ranking correlation is not only very strong but also highly statistically significant. 
Next, It is noted that since Sp cannot completely reach 1.0, therefore, the errors are inevitably present.  

\begin{figure}[t]
  \centering
  \vspace{-1pt}
\includegraphics[width=0.99\linewidth]{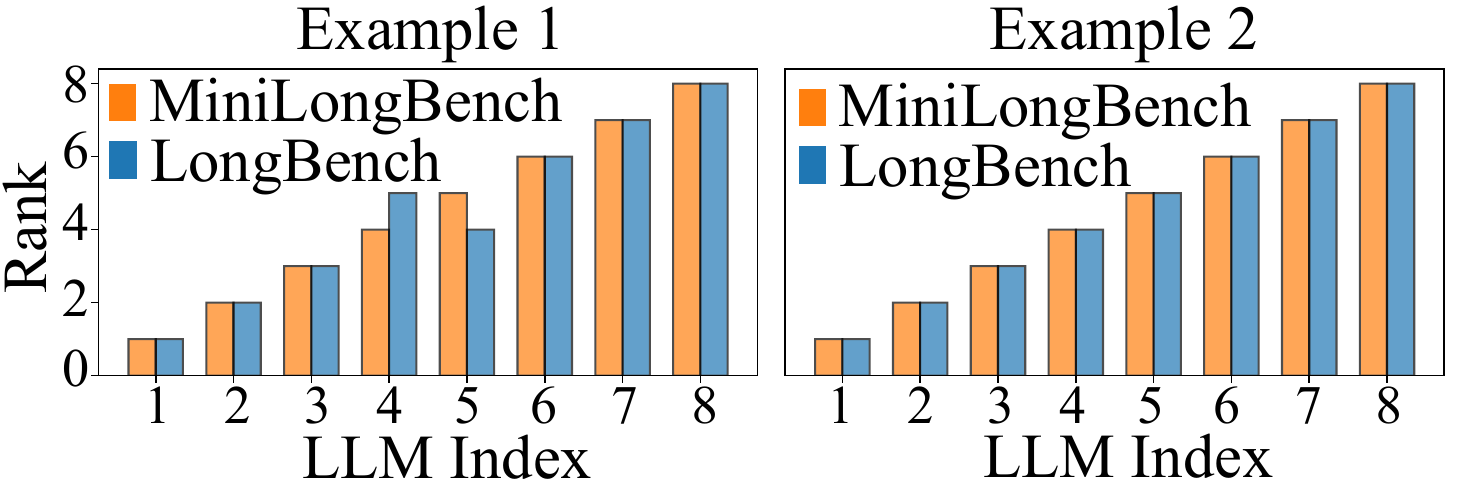}
\vspace{-15pt}
  \caption{The visualization of ranking. See more ranking examples in Appendix \ref{app:more}.}
  \vspace{-10pt}
\label{fig:rank}
\end{figure}

To demonstrate the usability of MiniLongBench with Sp = 0.97, in addition to the experiment in Fig.~\ref{fig:spall}, we consider directly visualizing the ranking results of different LCU benchmarks. As shown in Fig.~\ref{fig:rank}, for some random test samples, we also randomly selected 8 different LLMs to compare their ranking results on MiniLongBench and LongBench. We can observe that the ranking results are quite similar, despite some minor discrepancies. For more results, please refer to Appendix \ref{app:more}. In the future, we should further refine the compression methods to bring Sp as close as possible to 1.0 across all subtasks.

\begin{mdframed}[backgroundcolor=gray!8]
\begin{minipage}{\linewidth}
\vspace{2pt}
\noindent\textbf{(9) Why not just random sampling? }
\vspace{2pt}
\end{minipage}
\end{mdframed}
In Fig.~\ref{fig:redun}, we show that through random sampling, we identify a significant amount of redundancy in LongBench. However, relying solely on random sampling to compress LongBench is insufficient. The primary reason is that while random sampling can probabilistically yield high Sp results, as shown in Fig.~\ref{fig:redun}, the variance is substantial, making it easy to achieve suboptimal compression. 

The compression method we propose in Section \ref{sec:method} for the LCU benchmark effectively mitigates these issues, consistently achieving high Sp across various subtasks.

\section{Conclusion}

In this paper, we propose a concise data compression method for long-text data with sparse information. By pruning the well-known LCU benchmark LongBench, we created MiniLongBench. Through empirical analysis of over 60 LLMs with varying performance levels, MiniLongBench achieved an average evaluation cost reduction to 4.5\% of the original, while maintaining strong consistency with LongBench results. This phenomenon indicates that the proposed MiniLongBench has great potential to greatly promote the exploration of LLMs' LCU capabilities in the future.

\clearpage

\section*{Limitations}
The LCU benchmark compression method shown in this paper requires performance records from various LLMs as training data. However, most of this data is not open-source in practice. Consequently, we need to incur significant API costs and GPU computational resources to obtain this data. On the other hand, although we have achieved effective compression for LongBench, since Sp cannot be 1.0, we cannot expect MiniLongBench to have exactly the same evaluation capabilities as LongBench, only nearly identical. Additionally, there is still considerable room for performance improvement in the summarization and synthetic tasks, which are worthwhile directions for future enhancements.

\section*{Acknowledgments}
This work was supported by National Science and Technology Major Project (No.2021ZD0111601), National Natural Science Foundation of China under Grants No. 623B2099, 62272494 and 62325605, Guangdong Basic and Applied Basic Research Foundation (No.2023A1515011374, 2023A1515012845), and Guangzhou Science and Technology Program (No.2024A04J6365).

\bibliography{custom}

\clearpage
\appendix

\section{The Details of LongBench}
\label{app:LongBench}

LongBench~\citep{bai2023LongBench} represents the first bilingual, multi-task benchmark specifically developed for assessing long-context comprehension. The benchmark encompasses six primary task categories and 21 distinct tasks, spanning crucial long-text application domains ~\citep{dasigi2021dataset,yang2018hotpotqa,ho2020constructing,trivedi2022musique,huang2021efficient,zhong2021qmsum,fabbri2019multi,ainslie2023colt5,li2002learning} including multi-document QA, single-document QA, summarization, few-shot learning, code completion, and synthetic tasks, as detailed in Table~\ref{tb:stat}.

To thoroughly evaluate large models' bilingual proficiency in long-context processing, LongBench incorporates tasks in both Chinese and English. The dataset comprises 4,750 test instances, with average lengths of 6,711 words and 13,386 characters for English and Chinese respectively, ensuring extensive coverage of diverse scenarios.
The challenge of long-context understanding~\citep{press2022train,sun2022length,chen2023extending,zhong2024moextend,zhong2022cem} can be formally defined as follows: given an input sequence \( I \) and a context sequence \( C \), the model is tasked with generating an output \( A \). For example, in a QA task, \( I \) represents the question, \( C \) corresponds to the document, and \( A \) is the answer. Across LongBench, \( I \) and \( A \) are typically short, whereas \( C \) can span thousands of tokens. Specific instantiations of \((I, C, A)\) for each task are provided in Table~7 of \cite{bai2023LongBench}.

\section{The LLMs Considered in MiniLongBench}
\label{app:llms}
In this section, we list all LLMs we considered in Table \ref{tab:llmconsider}. Among them, 20 LLMs were utilized for training to aid in obtaining effective representations of test samples in the LCU benchmark. In this study, we have carefully curated a selection of LLMs that demonstrate consistently strong performance across multiple dimensions to contribute to the development of MiniLongBench. These models were chosen based on their proven capabilities in various tasks and benchmarks. However, to enhance the scalability and objectivity of our approach, future work should focus on implementing an automated LLM selection mechanism. This advancement would not only streamline the selection process but also ensure a more systematic and unbiased evaluation of potential models for inclusion in MiniLongBench.

% Table generated by Excel2LaTeX from sheet 'Sheet1'
\begin{table}[htbp]
  \centering
\resizebox*{0.99\linewidth}{!}{
    \begin{tabular}{ll|ll}
    \hline
    \textbf{Model} & \textbf{Type} & \textbf{Model} & \textbf{Type} \\
    \hline
    ALMA-7B-Ja-V2 & T    & Amd-llama-135m & A \\
    \rowcolor{mygray}GOAT-7B-Community & T    & Amd-llama-135m-code & A \\
    Koss-7B-chat & T    & Distilled-HermesChat-7B & A \\
    \rowcolor{mygray}Kunoichi-7B & T    & Loyal-Macaroni-Maid-7B & A \\
    Llama-2-7b-ft-instruct-es & T    & Llama-3-8b-hf & A \\
    \rowcolor{mygray}Llama-2-7b-hf & T    & Llama-7b-SFT\_ds\_wiki65k & A \\
    Llama-7b-hf & T    & Llama-shishya-7b-ep3-v1 & A \\
    \rowcolor{mygray}Mistral-7B-Instruct-v0.2 & T    & Llama-30b & A \\
    OLMo-1B & T    & OLMo-1B-SFT & A \\
    \rowcolor{mygray}Airoboros-7b & T    & StopCarbon-10.7B-v6 & A \\
    Gemma2-9b-hf & T    & Synatra-RP-Orca-2-7b-v0.1 & A \\
    \rowcolor{mygray}Giraffe-7b & T    & TowerInstruct-7B-v0.1 & A \\
    Mistral-7b-v0.1-hf & T    & Gemma2-2b-hf & A \\
    \rowcolor{mygray}Perry-7b & T    & Manatee-7b & A \\
    Qwen-7b-hf & T    & Mistral-7b-v0.3-hf & A \\
    \rowcolor{mygray}Qwen1.5-0.5b-hf & T  & Qwen1.5-1.8b-hf & A \\
    Qwen2-0.5b-hf & T  & Qwen2.5-0.5b-instruct-hf & A \\
    \rowcolor{mygray}Qwen2.5-0.5b-base & T  & Qwen2.5-3b-base & A \\
    Qwen2.5-1.5b-base & T  & Qwen2.5-3b-instruct-hf & A \\
    \rowcolor{mygray}Tulu-7B-fp16 & T    & Recycled-wizardlm-7b-v2.0 & A \\
    OPT-30B & A  & Wizaed-Vicuna & A \\
    \rowcolor{mygray}Llama-30B & A  & LwQ-Instruct & A \\
    30B-Epsilon & A  & DeepSeek V3 & A \\
    \rowcolor{mygray}GPT o1 mini & A    & GPT-3.5-turbo & A \\
    ChatGLM-6B & A  & ChatGLM3-6B & A \\
    \rowcolor{mygray}ChatGLM3-9B & A  & Qwen-7B & A \\
    Qwen2-7B & A  & Qwen2.5-7B & A \\
     \rowcolor{mygray}Qwen2.5-14B & A  &  Qwen2.5-32B & A \\
    Gemma2-9B & A  & Llama-7B-2k & A \\
     \rowcolor{mygray}Llama2-7B-4k & A  &  Llama3-8B & A \\
    \hline
    \end{tabular}%
    }
      \caption{The LLMs considered in MiniLongBench. "T" and "A" denote "for tranining" and "for analysis".}
  \label{tab:llmconsider}%
\end{table}%

\section{The Details Results of Advanced LLMs}
\label{app:fur}
In Section \ref{app:perfo}, we present the direct performance results of some Advanced LLMs on the six main tasks of MiniLongBench. Note that MiniLongBench includes not only the six main tasks but also 21 subtasks. Therefore, in this section, we will display the detailed results. The results are shown in Table \ref{tb:exp1} and Table \ref{tb:exp2}, where the indices in the table correspond to those in Table \ref{tab:evalstat} in the main text.

In addition to the performance estimation of the target LLM on the entire LongBench using MiniLongBench's test samples, as demonstrated in Section \ref{app:perfo}, we further propose a more straightforward but slightly less effective method for evaluating LCU capabilities in Appendix \ref{app:dire}. Specifically, the target LLM is directly tested on MiniLongBench's test samples without requiring any additional steps.

% Table generated by Excel2LaTeX from sheet 'Sheet1'
\begin{table*}[t]
  \centering
  \resizebox*{0.99\linewidth}{!}{
    \begin{tabular}{l|ccccc|ccccc|ccccc}
    \hline
    \textbf{Model} & \multicolumn{5}{c|}{\textbf{Single-Doc QA}}    & \multicolumn{5}{c|}{\textbf{Multi-Doc QA} }    & \multicolumn{5}{c}{\textbf{Summarization}} \\
\cmidrule{2-16}          & \textbf{1-1}   & \textbf{1-2}   & \textbf{1-3 }  & \textbf{1-4}   & \textbf{Avg}   &\textbf{ 2-1}   & \textbf{2-2}   & \textbf{2-3}   & \textbf{2-4}   & \textbf{Avg}   & \textbf{3-1}   & \textbf{3-2}   & \textbf{3-3}   & \textbf{3-4}   & \textbf{Avg} \\
    \hline
   \rowcolor{mygray}DeepSeek-V3-128k & \textcolor[rgb]{ .094,  .102,  .106}{0.29 } & 0.43  & 0.50  & 0.52  & 0.43  & \textcolor[rgb]{ .094,  .102,  .106}{0.51 } & 0.45  & 0.21  & 0.10  & 0.31  & \textcolor[rgb]{ .094,  .102,  .106}{0.24 } & 0.08  & 0.16  & 0.01  & 0.12  \\
    GPT-4o-mini-128k & \textcolor[rgb]{ .094,  .102,  .106}{0.27 } & 0.41  & 0.49  & 0.48  & 0.41  & \textcolor[rgb]{ .094,  .102,  .106}{0.48 } & 0.42  & 0.19  & 0.09  & 0.30  & \textcolor[rgb]{ .094,  .102,  .106}{0.19 } & 0.07  & 0.15  & 0.01  & 0.10  \\
    \rowcolor{mygray}GPT-3.5-Turbo-16k & \textcolor[rgb]{ .094,  .102,  .106}{0.21 } & 0.36  & 0.47  & 0.43  & 0.37  & \textcolor[rgb]{ .094,  .102,  .106}{0.42 } & 0.37  & 0.15  & 0.07  & 0.25  & \textcolor[rgb]{ .094,  .102,  .106}{0.13 } & 0.04  & 0.10  & 0.01  & 0.07  \\
    Internlm3-8B-32k & \textcolor[rgb]{ .094,  .102,  .106}{0.17 } & 0.32  & 0.43  & 0.40  & 0.33  & \textcolor[rgb]{ .094,  .102,  .106}{0.40 } & 0.39  & 0.14  & 0.05  & 0.24  & \textcolor[rgb]{ .094,  .102,  .106}{0.07 } & 0.02  & 0.06  & 0.01  & 0.04  \\
    \rowcolor{mygray}ChatGLM3-6B-8k & \textcolor[rgb]{ .094,  .102,  .106}{0.05 } & 0.14  & 0.28  & 0.21  & 0.17  & \textcolor[rgb]{ .094,  .102,  .106}{0.12 } & 0.14  & 0.03  & 0.03  & 0.08  & \textcolor[rgb]{ .094,  .102,  .106}{0.03 } & 0.01  & 0.03  & 0.01  & 0.02  \\
    ChatGLM4-9B-128k & \textcolor[rgb]{ .094,  .102,  .106}{0.21 } & 0.36  & 0.44  & 0.45  & 0.36  & \textcolor[rgb]{ .094,  .102,  .106}{0.44 } & 0.42  & 0.17  & 0.06  & 0.27  & \textcolor[rgb]{ .094,  .102,  .106}{0.10 } & 0.04  & 0.09  & 0.01  & 0.06  \\
    \rowcolor{mygray}Qwen-7B-8k & \textcolor[rgb]{ .094,  .102,  .106}{0.08 } & 0.18  & 0.30  & 0.24  & 0.20  & \textcolor[rgb]{ .094,  .102,  .106}{0.17 } & 0.15  & 0.05  & 0.04  & 0.10  & \textcolor[rgb]{ .094,  .102,  .106}{0.07 } & 0.02  & 0.06  & 0.00  & 0.04  \\
    Qwen2-7B-128k & \textcolor[rgb]{ .094,  .102,  .106}{0.15 } & 0.30  & 0.38  & 0.34  & 0.29  & \textcolor[rgb]{ .094,  .102,  .106}{0.37 } & 0.36  & 0.12  & 0.05  & 0.22  & \textcolor[rgb]{ .094,  .102,  .106}{0.08 } & 0.03  & 0.07  & 0.01  & 0.05  \\
    \rowcolor{mygray}Qwen2.5-7B-128k & \textcolor[rgb]{ .094,  .102,  .106}{0.18 } & 0.32  & 0.43  & 0.41  & 0.34  & \textcolor[rgb]{ .094,  .102,  .106}{0.40 } & 0.38  & 0.14  & 0.05  & 0.24  & \textcolor[rgb]{ .094,  .102,  .106}{0.08 } & 0.03  & 0.07  & 0.01  & 0.04  \\
    Qwen2.5-14B-128k & \textcolor[rgb]{ .094,  .102,  .106}{0.19 } & 0.33  & 0.43  & 0.43  & 0.35  & \textcolor[rgb]{ .094,  .102,  .106}{0.42 } & 0.41  & 0.15  & 0.06  & 0.26  & \textcolor[rgb]{ .094,  .102,  .106}{0.08 } & 0.03  & 0.07  & 0.01  & 0.05  \\
    \rowcolor{mygray}Qwen2.5-32B-128k & \textcolor[rgb]{ .094,  .102,  .106}{0.21 } & 0.36  & 0.43  & 0.46  & 0.36  & \textcolor[rgb]{ .094,  .102,  .106}{0.43 } & 0.40  & 0.16  & 0.06  & 0.26  & \textcolor[rgb]{ .094,  .102,  .106}{0.11 } & 0.03  & 0.09  & 0.01  & 0.06  \\
    Llama-7B-2k & \textcolor[rgb]{ .094,  .102,  .106}{0.03 } & 0.07  & 0.17  & 0.08  & 0.09  & \textcolor[rgb]{ .094,  .102,  .106}{0.06 } & 0.05  & 0.02  & 0.02  & 0.04  & \textcolor[rgb]{ .094,  .102,  .106}{0.03 } & 0.01  & 0.04  & 0.00  & 0.02  \\
    \rowcolor{mygray}Llama2-7B-4k & \textcolor[rgb]{ .094,  .102,  .106}{0.05 } & 0.10  & 0.20  & 0.11  & 0.11  & \textcolor[rgb]{ .094,  .102,  .106}{0.06 } & 0.06  & 0.02  & 0.02  & 0.04  & \textcolor[rgb]{ .094,  .102,  .106}{0.03 } & 0.01  & 0.04  & 0.00  & 0.02  \\
    Llama3-8B-8k & \textcolor[rgb]{ .094,  .102,  .106}{0.03 } & 0.10  & 0.17  & 0.13  & 0.11  & \textcolor[rgb]{ .094,  .102,  .106}{0.03 } & 0.05  & 0.01  & 0.03  & 0.03  & \textcolor[rgb]{ .094,  .102,  .106}{0.09 } & 0.02  & 0.04  & 0.00  & 0.04  \\
    \rowcolor{mygray}Llama-30B-2k & \textcolor[rgb]{ .094,  .102,  .106}{0.04 } & 0.09  & 0.18  & 0.11  & 0.10  & \textcolor[rgb]{ .094,  .102,  .106}{0.05 } & 0.05  & 0.02  & 0.02  & 0.04  & \textcolor[rgb]{ .094,  .102,  .106}{0.05 } & 0.01  & 0.04  & 0.00  & 0.02  \\
    OPT-30B-2k & \textcolor[rgb]{ .094,  .102,  .106}{0.02 } & 0.07  & 0.16  & 0.08  & 0.08  & \textcolor[rgb]{ .094,  .102,  .106}{0.04 } & 0.05  & 0.01  & 0.02  & 0.03  & \textcolor[rgb]{ .094,  .102,  .106}{0.03 } & 0.00  & 0.02  & 0.00  & 0.01  \\
    \rowcolor{mygray}Wizard-Vicuna-2k & \textcolor[rgb]{ .094,  .102,  .106}{0.10 } & 0.18  & 0.33  & 0.17  & 0.20  & \textcolor[rgb]{ .094,  .102,  .106}{0.24 } & 0.20  & 0.06  & 0.02  & 0.13  & \textcolor[rgb]{ .094,  .102,  .106}{0.04 } & 0.01  & 0.04  & 0.00  & 0.02  \\
    LwQ-Instruct-2k & \textcolor[rgb]{ .094,  .102,  .106}{0.16 } & 0.22  & 0.34  & 0.19  & 0.23  & \textcolor[rgb]{ .094,  .102,  .106}{0.34 } & 0.24  & 0.11  & 0.02  & 0.18  & \textcolor[rgb]{ .094,  .102,  .106}{0.06 } & 0.02  & 0.08  & 0.00  & 0.04  \\
    \rowcolor{mygray}30B-Epsilon-2k & \textcolor[rgb]{ .094,  .102,  .106}{0.13 } & 0.19  & 0.30  & 0.16  & 0.20  & \textcolor[rgb]{ .094,  .102,  .106}{0.24 } & 0.17  & 0.08  & 0.02  & 0.13  & \textcolor[rgb]{ .094,  .102,  .106}{0.05 } & 0.02  & 0.07  & 0.01  & 0.04  \\
    \hline
    \end{tabular}%
    }
\caption{Results on single-doc QA, multi-doc QA and summarization tasks. The indexes, like "1-1" or "4-1", are following Table \ref{tb:stat}. "avg" represents the average performance of subtasks under different main tasks.}
\label{tb:exp1}
\end{table*}%

\begin{figure*}[t]
  \centering
  \vspace{-5pt}
\includegraphics[width=0.99\linewidth]{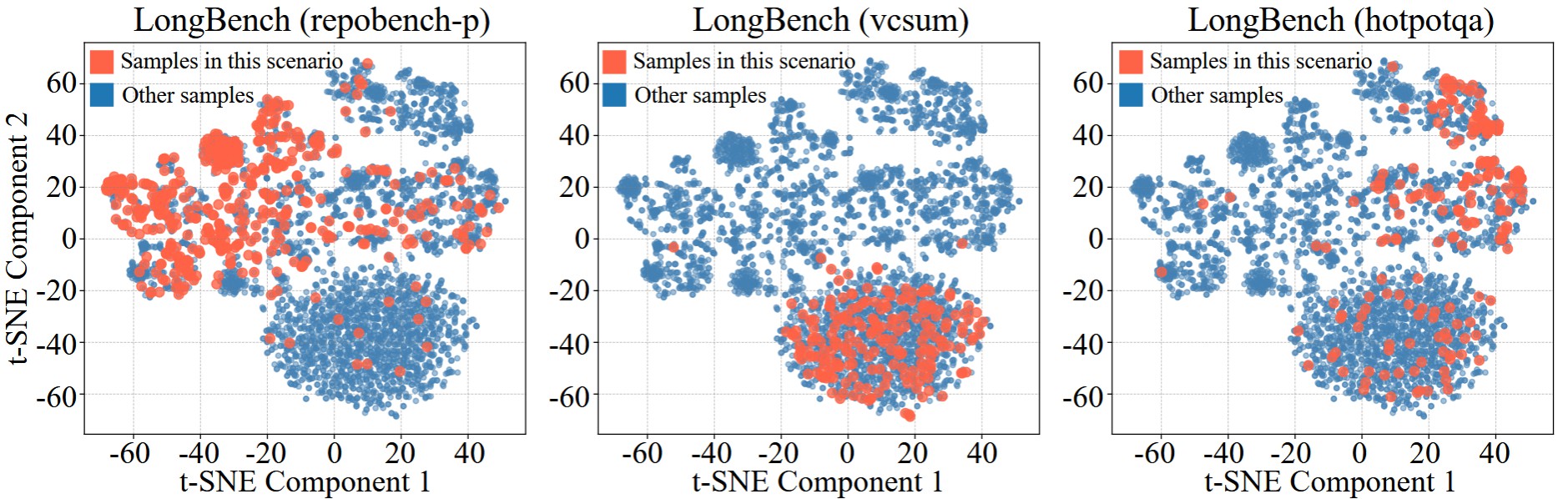}
\vspace{-5pt}
  \caption{The visualization of learned representation $(e_j,\beta_j)$ of test sample.}
  % \vspace{-10pt}
\label{fig:vis}
\end{figure*}
% Table generated by Excel2LaTeX from sheet 'Sheet1'
\begin{table*}[t]
  \centering
  \resizebox*{0.99\linewidth}{!}{
    \begin{tabular}{l|ccccc|cccc|ccc|ccc}
    \hline
    \textbf{Model} & \multicolumn{5}{c|}{\textbf{Few-show Learning}} & \multicolumn{4}{c|}{\textbf{Synthetic}} & \multicolumn{3}{c|}{\textbf{Code}} & \multicolumn{3}{c}{\textbf{Overall}} \\
\cmidrule{2-16}          & \textbf{4-1}   & \textbf{4-2}   & \textbf{4-3 }  & \textbf{4-4 }  & \textbf{Avg}   & \textbf{5-1}   & \textbf{5-2}   & \textbf{5-3}   & \textbf{Avg}   & \textbf{6-1}   & \textbf{6-2}   & \textbf{Avg}   & \textbf{EN}    & \textbf{ZH}    & \textbf{All} \\
    \hline
    \rowcolor{mygray}DeepSeek-V3-128k & 0.77  & 0.93  & 0.63  & 0.35  & 0.67  & 0.05  & 0.38  & 0.34  & 0.26  & 0.90  & 0.86  & 0.88  & 0.43  & 0.26  & 0.45  \\
    GPT-4o-mini-128k & 0.76  & 0.93  & 0.60  & 0.32  & 0.65  & 0.04  & 0.35  & 0.28  & 0.23  & 0.90  & 0.85  & 0.87  & 0.42  & 0.24  & 0.43  \\
    \rowcolor{mygray}GPT-3.5-Turbo-16k & 0.70  & 0.88  & 0.49  & 0.23  & 0.58  & 0.03  & 0.27  & 0.21  & 0.17  & 0.84  & 0.78  & 0.81  & 0.37  & 0.19  & 0.37  \\
    Internlm3-8B-32k & 0.61  & 0.86  & 0.48  & 0.17  & 0.53  & 0.01  & 0.18  & 0.16  & 0.12  & 0.67  & 0.62  & 0.64  & 0.32  & 0.16  & 0.32  \\
    \rowcolor{mygray}ChatGLM3-6B-8k & 0.47  & 0.69  & 0.26  & 0.13  & 0.39  & 0.00  & 0.03  & 0.04  & 0.02  & 0.49  & 0.48  & 0.48  & 0.19  & 0.09  & 0.19  \\
    ChatGLM4-9B-128k & 0.69  & 0.89  & 0.51  & 0.24  & 0.58  & 0.02  & 0.24  & 0.16  & 0.14  & 0.79  & 0.70  & 0.75  & 0.36  & 0.18  & 0.36  \\
    \rowcolor{mygray}Qwen-7B-8k & 0.58  & 0.78  & 0.34  & 0.20  & 0.48  & 0.01  & 0.12  & 0.08  & 0.07  & 0.74  & 0.68  & 0.71  & 0.25  & 0.11  & 0.27  \\
    Qwen2-7B-128k & 0.66  & 0.89  & 0.44  & 0.20  & 0.55  & 0.01  & 0.15  & 0.11  & 0.09  & 0.74  & 0.69  & 0.71  & 0.32  & 0.14  & 0.32  \\
    \rowcolor{mygray}Qwen2.5-7B-128k & 0.65  & 0.87  & 0.49  & 0.21  & 0.55  & 0.01  & 0.19  & 0.16  & 0.12  & 0.69  & 0.65  & 0.67  & 0.33  & 0.17  & 0.33  \\
    Qwen2.5-14B-128k & 0.65  & 0.87  & 0.49  & 0.22  & 0.56  & 0.01  & 0.19  & 0.16  & 0.12  & 0.68  & 0.64  & 0.66  & 0.33  & 0.17  & 0.33  \\
    \rowcolor{mygray}Qwen2.5-32B-128k & 0.69  & 0.90  & 0.54  & 0.25  & 0.59  & 0.02  & 0.24  & 0.19  & 0.15  & 0.76  & 0.71  & 0.74  & 0.36  & 0.20  & 0.36  \\
    Llama-7B-2k & 0.47  & 0.71  & 0.21  & 0.15  & 0.39  & 0.00  & 0.04  & 0.03  & 0.02  & 0.64  & 0.58  & 0.61  & 0.18  & 0.06  & 0.20  \\
    \rowcolor{mygray}Llama2-7B-4k & 0.53  & 0.75  & 0.22  & 0.15  & 0.41  & 0.00  & 0.07  & 0.05  & 0.04  & 0.69  & 0.63  & 0.66  & 0.20  & 0.07  & 0.21  \\
    Llama3-8B-8k & 0.60  & 0.74  & 0.30  & 0.24  & 0.47  & 0.01  & 0.10  & 0.14  & 0.08  & 0.74  & 0.68  & 0.71  & 0.22  & 0.11  & 0.24  \\
    \rowcolor{mygray}Llama-30B-2k & 0.50  & 0.72  & 0.24  & 0.17  & 0.41  & 0.00  & 0.08  & 0.05  & 0.04  & 0.67  & 0.62  & 0.65  & 0.20  & 0.07  & 0.21  \\
    OPT-30B-2k & 0.40  & 0.65  & 0.16  & 0.11  & 0.33  & 0.00  & 0.02  & 0.03  & 0.02  & 0.49  & 0.47  & 0.48  & 0.15  & 0.05  & 0.16  \\
    \rowcolor{mygray}Wizard-Vicuna-2k & 0.52  & 0.76  & 0.24  & 0.13  & 0.41  & 0.01  & 0.08  & 0.05  & 0.05  & 0.62  & 0.55  & 0.59  & 0.23  & 0.08  & 0.23  \\
    LwQ-Instruct-2k & 0.54  & 0.78  & 0.30  & 0.16  & 0.45  & 0.01  & 0.13  & 0.07  & 0.07  & 0.73  & 0.66  & 0.70  & 0.28  & 0.09  & 0.28  \\
    \rowcolor{mygray}30B-Epsilon-2k & 0.58  & 0.83  & 0.34  & 0.18  & 0.48  & 0.01  & 0.13  & 0.07  & 0.07  & 0.76  & 0.70  & 0.73  & 0.27  & 0.09  & 0.27  \\
    \hline
    \end{tabular}%
    }
\caption{Results on few-shot learning, synthetic, and code tasks. `Overall' is computed by the macro-average (the mean of `Avg') over major task categories. This is computed on English (EN) tasks, Chinese (ZH) tasks, and all (All) tasks, code tasks are included in both languages. The indexes, like "1-1" or "4-1", are following Table \ref{tb:stat}. "avg" represents the average performance of subtasks under different main tasks.}
\label{tb:exp2}
\end{table*}%

%\rowcolor{mygray}

\begin{figure*}[t]
  \centering
  \vspace{-5pt}
\includegraphics[width=0.96\linewidth]{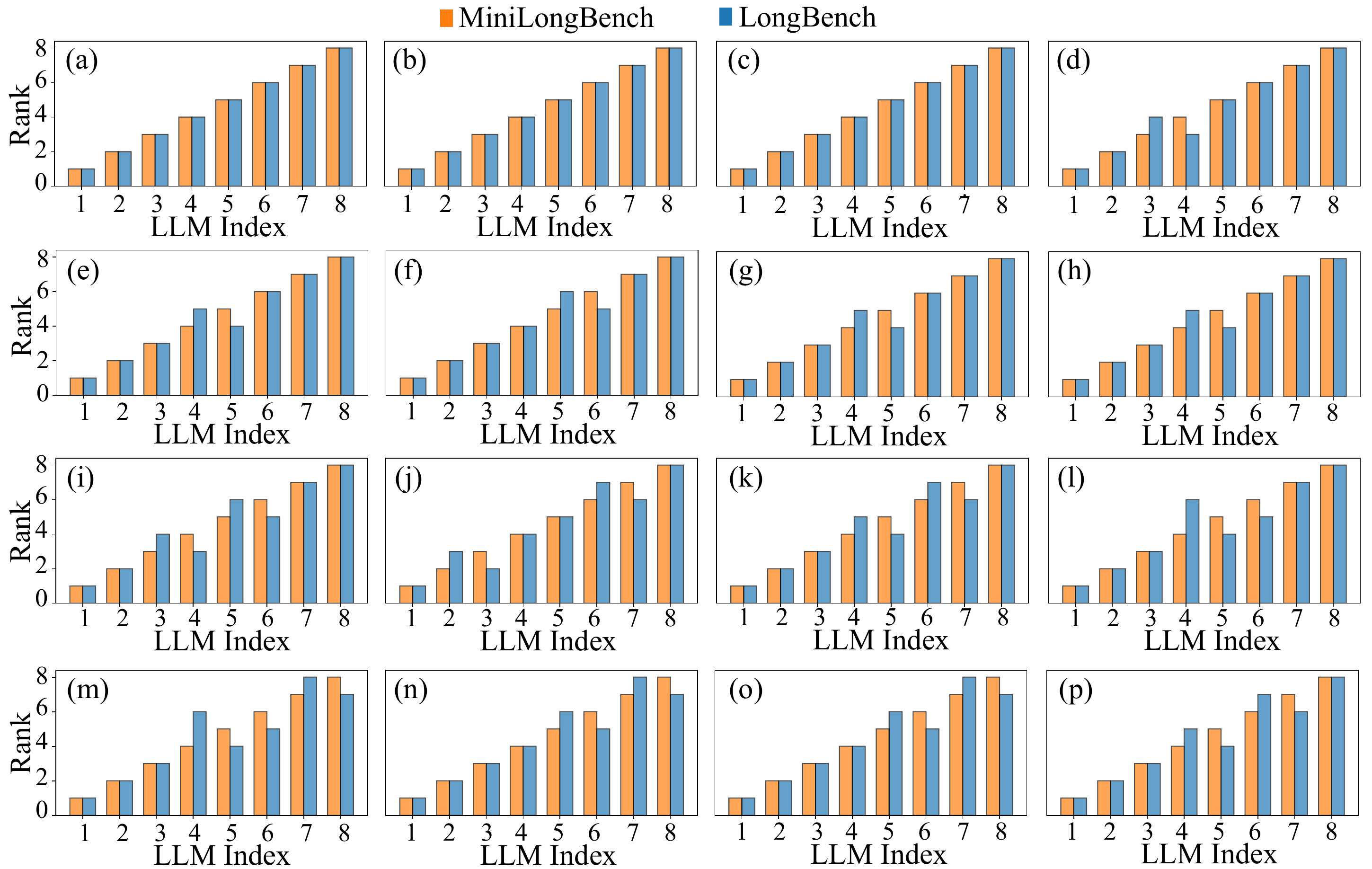}
\vspace{-5pt}
  \caption{The more examples of  visualization of ranking by MiniLongBench and LongBench..}
  % \vspace{-10pt}
\label{fig:exa}
\end{figure*}

\section{Related Works}
\label{related}
\subsection{Long Context Understanding (LCU)}
% \noindent\textbf{Long Context Understanding (LCU)}.
Existing research on LCU in LLMs primarily addresses two critical challenges in long-text modeling: the substantial runtime overhead associated with extended contexts and the issue of catastrophic forgetting during long sequence processing. 

A significant body of work has concentrated on enhancing the efficiency and memory retention of Transformers~\citep{tay2022efficient}. This includes innovations in sparse and efficient computation~\citep{child2019generating,kitaev2020reformer,beltagy2020longformer,zaheer2020big,wang2020linformer,fedus2022switch,ding2023longnet}, as well as the integration of recurrent and memory modules~\citep{dai2019transformer,rae2020compressive,wu2022memorizing,martins2022former,bulatov2022recurrent,orvieto2023resurrecting,liang2023unleashing,zhou2023recurrentgpt}.

More recently, several advanced methods~\citep{press2022train,sun2022length,chen2023extending} have been developed to facilitate length extrapolation in Transformers. These techniques have been incorporated into the training frameworks of long-context LLMs such as ChatGLM2-32k~\citep{zeng2022glm} and LongChat-32k~\citep{longchat2023}, among others. These models have successfully extended their context lengths to 128k tokens or more~\cite{claude-3-5,GPT-4o,reid2024gemini,glm2024chatglm,dubey2024llama,xiong2024effective,pmlr-v235-fu24d,bai2024longalign,gao2024train}, marking a significant advancement in the field.

\subsection{The LCU Benchmarks for LLMs}
% \noindent\textbf{The LCU benchmarks for LLMs}.
Given the critical importance of LCU capabilities for LLMs, an increasing number of benchmarks have been proposed to evaluate these capabilities, playing a pivotal role in exploring and advancing LLMs' LCU proficiency. A significant portion of these benchmarks of LLMs focuses on comprehensive LCU assessment, encompassing tasks such as Question Answering, information retrieval, and summarization. 
Notable examples include L-Eval~\cite{an2024leval}, LongBench~\cite{bai2024LongBench}, ZeroSCROLLS~\cite{shaham2023zeroscrolls}, BAMBOO~\cite{dong2024bamboo}, LooGLE~\cite{li2023loogle}, $\infty$-bench~\cite{zhang2024infty}, Ruler~\cite{hsieh2024ruler}, and HELMET~\cite{yen2024helmet}.
Another category of benchmarks is specifically designed to explore particular aspects of LCU capabilities. These include retrieval and attribution tasks~\cite{needleinhaystack,kuratov2024babilong,song2024counting,laban2024summary,zhang2024longcite,vodrahalli2024michelangelo,krishna2024fact}, document QA~\cite{dua2019drop,dasigi2021dataset,pang2022quality,wang2024leave}, summarization~\cite{zhong2021qmsum,huang2021efficient,wang2022squality}, coding~\cite{liu2023repobench,bogomolov2024long}, many-shot learning~\cite{agarwal2024many}, and long-text generation~\cite{bai2024longwriter,wu2024longgenbench,liu2024longgenbench,que2024hellobench}. 

These specialized benchmarks provide targeted insights into the diverse and complex facets of LCU, contributing to a more nuanced understanding and development of LLMs' long-context processing abilities.

\subsection{Low-cost Deep Learning}
% \noindent\textbf{Low-cost Deep Learning}.
Recently, there has been a surge of efforts aimed at achieving low-cost deep learning, encompassing strategies such as the compression of model parameters or the design of lightweight architectures~\citep{yang2024laco,muralidharan2024compact,lin2024mope,kim2024shortened,zhong2023lsas,he2021blending,huang2022lottery,huang2021rethinking,liang2020instance}.
Concurrently, some research has explored compressing the training dataset~\citep{gadre2024datacomp,sachdeva2023data,yu2023dataset,lei2023comprehensive,touvron2021training}
to reduce computational costs while maintaining performance. Beyond these approaches, in the era of large language models, works including this paper consider compressing test data~\citep{polotinybenchmarks,pacchiardi2024100,kipnis2024texttt} as an effective means to aid in model architecture design, parameter tuning, and other training-related processes, thereby accelerating the iteration speed of robust models.

\section{The Visualization of Learned Representation of Test Samples}
\label{app:bench}
%LongBench

\begin{figure*}[t]
  \centering
  \vspace{-5pt}
\includegraphics[width=0.95\linewidth]{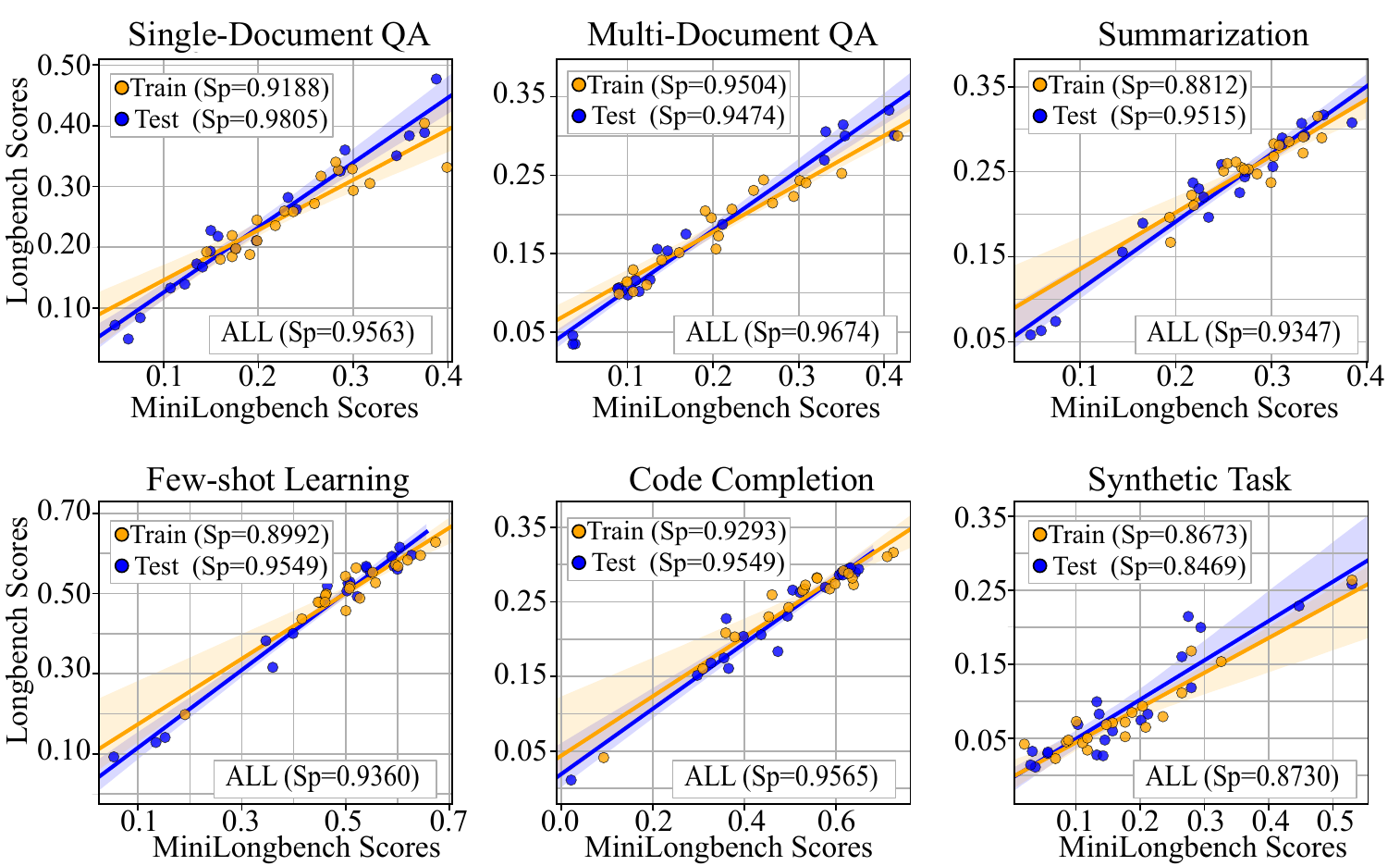}
\vspace{-3pt}
  \caption{The analysis of rank correlation (Sp) between LongBench and MiniLongBench where the result of MiniLongBench is evaluating directly.}
  \vspace{-10pt}
\label{fig:spall2}
\end{figure*}

In Section \ref{sec:method}, we use the performance record of $m$ LLMs on various test samples in LongBench, and use a logistic regression model for representation learning, obtaining their representations $(e_j, \beta_j)$. In Fig.~\ref{fig:vis}, we visualize test samples from certain sub-tasks listed in Table \ref{tb:stat} using t-SNE~\citep{van2008visualizing}. It can be observed that many test samples form clusters, and the representations of samples within the same cluster are highly similar. This further demonstrates that LongBench contains a significant amount of redundancy in its data, and the representation learning method proposed in Section \ref{sec:method} is effective for identifying redundant data in LongBench through clustering.

\section{More Visualizations of Ranking}
\label{app:more}
In the Fig.~\ref{fig:rank}, we provided some examples of rankings by MiniLongBench and LongBench. In this section, we will present more random examples to illustrate the usability and reliability of MiniLongBench. The results are shown in Fig.~\ref{fig:exa}. 
As illustrated in Fig.~\ref{fig:exa}, similar to the observations in the main text's Fig.~\ref{fig:rank}, the results from 16 random sampling trials consistently demonstrate that the ranking outcomes of various LLMs on MiniLongBench closely align with those on LongBench. Although minor discrepancies exist, they are within an acceptable range, particularly considering that the Spearman correlation coefficient (Sp) does not reach a perfect 1.0. These visualizations further validate that MiniLongBench achieves evaluation results comparable to LongBench while significantly reducing computational costs. This highlights MiniLongBench's effectiveness as a low-cost alternative for assessing LLM performance.

\begin{table}[t]
  \centering
  
  \resizebox*{0.99\linewidth}{!}{
    \begin{tabular}{l|cccccc}
    \toprule
    \textbf{Model} & \textbf{SQA}   & \textbf{MQA}   & \textbf{SUM}   & \textbf{FSL}   & \textbf{SYN}   & \textbf{CODE} \\
    \midrule
    \rowcolor{mygray}DeepSeek-V3-128k & 0.58  & 0.61  & 0.17  & 0.72  & 0.67  & 0.72  \\
    GPT-4o-mini-128k & 0.54  & 0.57  & 0.18  & 0.61  & 0.67  & 0.70  \\
    \rowcolor{mygray}GPT-3.5-Turbo-16k & 0.43  & 0.47  & 0.17  & 0.40  & 0.64  & 0.61  \\
    Internlm3-8B-32k & 0.38  & 0.55  & 0.16  & 0.34  & 0.67  & 0.19  \\
    \rowcolor{mygray}ChatGLM3-6B-8k & 0.31  & 0.15  & 0.17  & 0.30  & 0.07  & 0.36  \\
    ChatGLM4-9B-128k & 0.48  & 0.56  & 0.18  & 0.56  & 0.58  & 0.36  \\
    \rowcolor{mygray}Qwen-7B-8k & 0.30  & 0.22  & 0.14  & 0.34  & 0.20  & 0.65  \\
    Qwen2-7B-128k & 0.42  & 0.42  & 0.16  & 0.49  & 0.36  & 0.43  \\
    \rowcolor{mygray}Qwen2.5-7B-128k & 0.48  & 0.50  & 0.17  & 0.60  & 0.67  & 0.10  \\
    Qwen2.5-14B-128k & 0.46  & 0.56  & 0.18  & 0.65  & 0.67  & 0.01  \\
    \rowcolor{mygray}Qwen2.5-32B-128k & 0.50  & 0.53  & 0.17  & 0.71  & 0.67  & 0.18  \\
    Llama-7B-2k & 0.13  & 0.09  & 0.12  & 0.25  & 0.05  & 0.57  \\
    \rowcolor{mygray}Llama2-7B-4k & 0.16  & 0.10  & 0.10  & 0.23  & 0.14  & 0.59  \\
    Llama3-8B-8k & 0.15  & 0.08  & 0.10  & 0.35  & 0.21  & 0.70  \\
    \rowcolor{mygray}Llama-30B-2k & 0.15  & 0.08  & 0.09  & 0.25  & 0.15  & 0.64  \\
    OPT-30B-2k & 0.11  & 0.09  & 0.11  & 0.15  & 0.07  & 0.44  \\
    \rowcolor{mygray}Wizard-Vicuna-2k & 0.29  & 0.26  & 0.16  & 0.27  & 0.16  & 0.43  \\
    LwQ-Instruct-2k & 0.34  & 0.36  & 0.16  & 0.31  & 0.18  & 0.61  \\
    \rowcolor{mygray}30B-Epsilon-2k & 0.26  & 0.21  & 0.17  & 0.42  & 0.20  & 0.62  \\
    \bottomrule
    \end{tabular}%
    }
    \caption{Specific evaluation results on evaluating directly by MiniLongBench. Due to differences in evaluation methods, the score values presented in this table vary somewhat from those in Table \ref{tab:evalstat}, but they yield a similar ranking of LLMs in terms of LCU capability.}
    \vspace{-5pt}
  \label{tab:evalstat22}%
\end{table}%

\section{Evaluating Directly by MiniLongBench}
\label{app:dire}
In Section \ref{app:perfo2} of the main text, we primarily introduce a method that utilizes test samples from MiniLongBench to assist in evaluating the performance of a target LLM on LongBench. This method achieves a performance of up to 0.97 in Sp. In practice, we can also directly test the target LLM on MiniLongBench's test samples to obtain an assessment of its LCU capability. The results in Fig.~\ref{fig:spall2} confirm that this evaluation method achieves good Sp across various tasks in MiniLongBench, with an average Sp of 0.95, slightly lower than the evaluation method presented in Section \ref{app:perfo2}. Furthermore, in Table \ref{tab:evalstat22}, we present the results of this evaluation method across six main tasks. Additionally, we provide more detailed results for each subtask in Table \ref{tb22:exp1} and Table \ref{tb22:exp2}.

It is noteworthy that, in practice, whether directly evaluating on LongBench or MiniLongBench, or using the predictive method in Section \ref{app:perfo2}, there may be some discrepancies in the score values. However, these discrepancies do not affect the ranking of LLMs' LCU capabilities. For instance, Fig.~\ref{fig:spall2} and the main text's Fig.~\ref{fig:spall} demonstrate that the results from different evaluation methods are highly consistent, despite minor deviations in score values. This phenomenon primarily arises from several factors: first, MiniLongBench involves significant pruning of test samples compared to LongBench, leading to unavoidable errors; second, during the logistic regression in Section \ref{sec:method}'s Eq.~(\ref{eq:logi}), normalization and discretization introduce certain errors, particularly in scaling. Fortunately, the primary goal of the LCU benchmark is to rank LLMs based on their LCU capabilities, so the absolute score values do not impact the final outcomes.

\section{The Cost by Fine-tuning $\bar{\theta}$}
\label{app:cost}
In Section \ref{app:perfo2}, additional fine-tuning of $\bar{\theta}$ is required, which primarily involves two costs: the training cost for fine-tuning and the storage cost for the representation vectors of LongBench's test samples. In practice, these costs are minimal and entirely acceptable. Specifically, storing the test samples of MiniLongBench and the representation vectors of LongBench's test samples requires only 9.01MB and 1.13MB of disk space, respectively. This is significantly lower and entirely acceptable compared to the original storage cost of nearly 200MB for LongBench's test samples. This reduction is largely due to the two-step dimensionality compression method described in Section \ref{sec:method}, which uses text embedding and PCA to compress each feature vector to a dimension of just $d=10$, thereby greatly reducing storage costs.

On the other hand, the cost of fine-tuning $\bar{\theta}$ is also very low and can even be performed on a standard laptop without the need for server-grade GPUs. This is because MiniLongBench contains only about 200 test samples, and the dimensions of all representation vectors are all $d=10$, so the logistic regression training does not require significant computational power. 
Through 100 repeated experiments, the average time required for fine-tuning $\bar{\theta}$ was calculated. on a server (CPU: AMD EPYC 7K62, GPU: RTX 3090 24GB) and a laptop (CPU: AMD Ryzen 6 5600H, GPU: RTX 3050 4GB), fine-tuning takes approximately 0.02 seconds and 0.03 seconds, respectively. Compared to the original testing time of LongBench shown in Fig.~\ref{fig:lowcost}, this is almost negligible.

% Table generated by Excel2LaTeX from sheet 'Sheet1'
\begin{table*}[t]
  \centering
  \resizebox*{0.99\linewidth}{!}{
    \begin{tabular}{l|ccccc|ccccc|ccccc}
    \hline
    \textbf{Model} & \multicolumn{5}{c|}{\textbf{Single-Doc QA}}    & \multicolumn{5}{c|}{\textbf{Multi-Doc QA} }    & \multicolumn{5}{c}{\textbf{Summarization}} \\
\cmidrule{2-16}          & \textbf{1-1}   & \textbf{1-2}   & \textbf{1-3 }  & \textbf{1-4}   & \textbf{Avg}   &\textbf{ 2-1}   & \textbf{2-2}   & \textbf{2-3}   & \textbf{2-4}   & \textbf{Avg}   & \textbf{3-1}   & \textbf{3-2}   & \textbf{3-3}   & \textbf{3-4}   & \textbf{Avg} \\
    \hline
    \rowcolor{mygray}DeepSeek-V3-128k & 0.66  & 0.40  & 0.53  & 0.72  & 0.58  & 0.74  & 0.89  & 0.59  & 0.22  & 0.61  & 0.18  & 0.21  & 0.16  & 0.15  & 0.17  \\
    GPT-4o-mini-128k & 0.51  & 0.47  & 0.52  & 0.67  & 0.54  & 0.64  & 0.81  & 0.54  & 0.29  & 0.57  & 0.16  & 0.20  & 0.17  & 0.18  & 0.18  \\
    \rowcolor{mygray}GPT-3.5-Turbo-16k & 0.12  & 0.53  & 0.48  & 0.60  & 0.43  & 0.51  & 0.72  & 0.44  & 0.20  & 0.47  & 0.14  & 0.20  & 0.17  & 0.17  & 0.17  \\
    Internlm3-8B-32k & 0.07  & 0.36  & 0.44  & 0.64  & 0.38  & 0.59  & 0.77  & 0.63  & 0.22  & 0.55  & 0.18  & 0.17  & 0.16  & 0.14  & 0.16  \\
    \rowcolor{mygray}ChatGLM3-6B-8k & 0.17  & 0.28  & 0.38  & 0.42  & 0.31  & 0.11  & 0.20  & 0.12  & 0.15  & 0.15  & 0.15  & 0.18  & 0.16  & 0.18  & 0.17  \\
    ChatGLM4-9B-128k & 0.33  & 0.40  & 0.56  & 0.62  & 0.48  & 0.63  & 0.76  & 0.67  & 0.20  & 0.56  & 0.16  & 0.21  & 0.16  & 0.19  & 0.18  \\
    \rowcolor{mygray}Qwen-7B-8k & 0.16  & 0.28  & 0.46  & 0.32  & 0.30  & 0.29  & 0.22  & 0.22  & 0.13  & 0.22  & 0.17  & 0.16  & 0.17  & 0.06  & 0.14  \\
    Qwen2-7B-128k & 0.38  & 0.40  & 0.38  & 0.52  & 0.42  & 0.64  & 0.71  & 0.19  & 0.15  & 0.42  & 0.18  & 0.18  & 0.15  & 0.13  & 0.16  \\
    \rowcolor{mygray}Qwen2.5-7B-128k & 0.50  & 0.37  & 0.46  & 0.58  & 0.48  & 0.71  & 0.66  & 0.50  & 0.14  & 0.50  & 0.18  & 0.21  & 0.14  & 0.15  & 0.17  \\
    Qwen2.5-14B-128k & 0.41  & 0.43  & 0.44  & 0.56  & 0.46  & 0.67  & 0.89  & 0.44  & 0.23  & 0.56  & 0.16  & 0.23  & 0.15  & 0.16  & 0.18  \\
    \rowcolor{mygray}Qwen2.5-32B-128k & 0.45  & 0.43  & 0.42  & 0.68  & 0.50  & 0.66  & 0.76  & 0.48  & 0.21  & 0.53  & 0.16  & 0.21  & 0.15  & 0.17  & 0.17  \\
    Llama-7B-2k & 0.03  & 0.11  & 0.23  & 0.12  & 0.13  & 0.07  & 0.08  & 0.15  & 0.04  & 0.09  & 0.14  & 0.05  & 0.18  & 0.12  & 0.12  \\
    \rowcolor{mygray}Llama2-7B-4k & 0.13  & 0.20  & 0.23  & 0.09  & 0.16  & 0.05  & 0.11  & 0.13  & 0.10  & 0.10  & 0.16  & 0.07  & 0.04  & 0.14  & 0.10  \\
    Llama3-8B-8k & 0.02  & 0.20  & 0.17  & 0.22  & 0.15  & 0.04  & 0.10  & 0.13  & 0.06  & 0.08  & 0.18  & 0.14  & 0.01  & 0.09  & 0.10  \\
    \rowcolor{mygray}Llama-30B-2k & 0.03  & 0.17  & 0.27  & 0.11  & 0.15  & 0.07  & 0.12  & 0.08  & 0.04  & 0.08  & 0.16  & 0.05  & 0.05  & 0.11  & 0.09  \\
    OPT-30B-2k & 0.03  & 0.18  & 0.11  & 0.11  & 0.11  & 0.06  & 0.09  & 0.14  & 0.07  & 0.09  & 0.18  & 0.05  & 0.14  & 0.08  & 0.11  \\
    \rowcolor{mygray}Wizard-Vicuna-2k & 0.28  & 0.21  & 0.49  & 0.16  & 0.29  & 0.46  & 0.26  & 0.19  & 0.13  & 0.26  & 0.16  & 0.16  & 0.18  & 0.14  & 0.16  \\
    LwQ-Instruct-2k & 0.33  & 0.36  & 0.46  & 0.20  & 0.34  & 0.58  & 0.40  & 0.37  & 0.09  & 0.36  & 0.17  & 0.18  & 0.19  & 0.10  & 0.16  \\
    \rowcolor{mygray}30B-Epsilon-2k & 0.24  & 0.21  & 0.40  & 0.18  & 0.26  & 0.27  & 0.22  & 0.28  & 0.07  & 0.21  & 0.14  & 0.18  & 0.18  & 0.17  & 0.17  \\
    \hline
    \end{tabular}%
    }
\caption{Results on single-doc QA, multi-doc QA and summarization tasks based on evaluating directly by MiniLongBench. The indexes, like "1-1" or "4-1", are following Table \ref{tb:stat}. "avg" represents the average performance of subtasks under different main tasks.}
\label{tb22:exp1}
\end{table*}%

% Table generated by Excel2LaTeX from sheet 'Sheet1'
\begin{table*}[t]
  \centering
  \resizebox*{0.99\linewidth}{!}{
    \begin{tabular}{l|ccccc|cccc|ccc|ccc}
    \hline
    \textbf{Model} & \multicolumn{5}{c|}{\textbf{Few-show Learning}} & \multicolumn{4}{c|}{\textbf{Synthetic}} & \multicolumn{3}{c|}{\textbf{Code}} & \multicolumn{3}{c}{\textbf{Overall}} \\
\cmidrule{2-16}          & \textbf{4-1}   & \textbf{4-2}   & \textbf{4-3 }  & \textbf{4-4 }  & \textbf{Avg}   & \textbf{5-1}   & \textbf{5-2}   & \textbf{5-3}   & \textbf{Avg}   & \textbf{6-1}   & \textbf{6-2}   & \textbf{Avg}   & \textbf{EN}    & \textbf{ZH}    & \textbf{All} \\
    \hline
    \rowcolor{mygray}DeepSeek-V3-128k & 0.63  & 1.00  & 0.38  & 0.88  & 0.72  & 0.00  & 1.00  & 1.00  & 0.67  & 0.83  & 0.62  & 0.72  & 0.52  & 0.59  & 0.58  \\
    GPT-4o-mini-128k & 0.63  & 1.00  & 0.31  & 0.50  & 0.61  & 0.00  & 1.00  & 1.00  & 0.67  & 0.80  & 0.61  & 0.70  & 0.49  & 0.53  & 0.54  \\
    \rowcolor{mygray}GPT-3.5-Turbo-16k & 0.63  & 0.60  & 0.38  & 0.00  & 0.40  & 0.00  & 0.93  & 1.00  & 0.64  & 0.73  & 0.49  & 0.61  & 0.42  & 0.39  & 0.45  \\
    Internlm3-8B-32k & 0.00  & 1.00  & 0.38  & 0.00  & 0.34  & 0.00  & 1.00  & 1.00  & 0.67  & 0.19  & 0.18  & 0.19  & 0.36  & 0.40  & 0.38  \\
    \rowcolor{mygray}ChatGLM3-6B-8k & 0.50  & 0.01  & 0.30  & 0.38  & 0.30  & 0.00  & 0.00  & 0.20  & 0.07  & 0.45  & 0.27  & 0.36  & 0.19  & 0.26  & 0.22  \\
    ChatGLM4-9B-128k & 0.63  & 0.83  & 0.35  & 0.44  & 0.56  & 0.00  & 1.00  & 0.73  & 0.58  & 0.59  & 0.13  & 0.36  & 0.44  & 0.44  & 0.45  \\
    \rowcolor{mygray}Qwen-7B-8k & 0.44  & 0.25  & 0.16  & 0.50  & 0.34  & 0.00  & 0.40  & 0.20  & 0.20  & 0.76  & 0.55  & 0.65  & 0.28  & 0.24  & 0.31  \\
    Qwen2-7B-128k & 0.38  & 0.96  & 0.35  & 0.27  & 0.49  & 0.00  & 0.47  & 0.60  & 0.36  & 0.54  & 0.32  & 0.43  & 0.37  & 0.34  & 0.38  \\
    \rowcolor{mygray}Qwen2.5-7B-128k & 0.50  & 0.83  & 0.37  & 0.69  & 0.60  & 0.00  & 1.00  & 1.00  & 0.67  & 0.10  & 0.11  & 0.10  & 0.39  & 0.51  & 0.42  \\
    Qwen2.5-14B-128k & 0.63  & 0.86  & 0.35  & 0.75  & 0.65  & 0.00  & 1.00  & 1.00  & 0.67  & 0.00  & 0.03  & 0.01  & 0.39  & 0.54  & 0.42  \\
    \rowcolor{mygray}Qwen2.5-32B-128k & 0.63  & 0.97  & 0.35  & 0.88  & 0.71  & 0.00  & 1.00  & 1.00  & 0.67  & 0.29  & 0.07  & 0.18  & 0.41  & 0.59  & 0.46  \\
    Llama-7B-2k & 0.31  & 0.31  & 0.12  & 0.25  & 0.25  & 0.00  & 0.07  & 0.07  & 0.05  & 0.71  & 0.44  & 0.57  & 0.18  & 0.12  & 0.20  \\
    \rowcolor{mygray}Llama2-7B-4k & 0.25  & 0.25  & 0.14  & 0.29  & 0.23  & 0.00  & 0.20  & 0.22  & 0.14  & 0.71  & 0.48  & 0.59  & 0.18  & 0.17  & 0.22  \\
    Llama3-8B-8k & 0.38  & 0.20  & 0.16  & 0.69  & 0.35  & 0.00  & 0.17  & 0.47  & 0.21  & 0.84  & 0.57  & 0.70  & 0.19  & 0.30  & 0.27  \\
    \rowcolor{mygray}Llama-30B-2k & 0.31  & 0.24  & 0.15  & 0.31  & 0.25  & 0.00  & 0.32  & 0.13  & 0.15  & 0.76  & 0.52  & 0.64  & 0.19  & 0.14  & 0.23  \\
    OPT-30B-2k & 0.17  & 0.21  & 0.11  & 0.13  & 0.15  & 0.02  & 0.10  & 0.09  & 0.07  & 0.51  & 0.37  & 0.44  & 0.14  & 0.09  & 0.16  \\
    \rowcolor{mygray}Wizard-Vicuna-2k & 0.25  & 0.42  & 0.17  & 0.23  & 0.27  & 0.00  & 0.27  & 0.20  & 0.16  & 0.51  & 0.35  & 0.43  & 0.26  & 0.17  & 0.26  \\
    LwQ-Instruct-2k & 0.31  & 0.31  & 0.17  & 0.44  & 0.31  & 0.00  & 0.33  & 0.20  & 0.18  & 0.71  & 0.52  & 0.61  & 0.32  & 0.21  & 0.33  \\
    \rowcolor{mygray}30B-Epsilon-2k & 0.24  & 0.80  & 0.26  & 0.38  & 0.42  & 0.00  & 0.40  & 0.20  & 0.20  & 0.75  & 0.50  & 0.62  & 0.30  & 0.20  & 0.31  \\
    \hline
    \end{tabular}%
    }
\caption{Results on few-shot learning, synthetic, and code tasks  based on evaluating directly by MiniLongBench. `Overall' is computed by the macro-average (the mean of `Avg') over major task categories. This is computed on English (EN) tasks, Chinese (ZH) tasks, and all (All) tasks, code tasks are included in both languages. The indexes, like "1-1" or "4-1", are following Table \ref{tb:stat}. "avg" represents the average performance of subtasks under different main tasks.}
\label{tb22:exp2}
\end{table*}%

\end{document}